\documentclass[10pt]{sensys-proc}
\pdfoutput=1

\usepackage{graphicx}
\usepackage{color}
\usepackage{subfig}
\usepackage{morefloats}
\usepackage{adjustbox}

\begin{document}

%
\title{Feasibility of Video-based Sub-meter Localization on Resource-constrained Platforms}

%

\numberofauthors{2}
\author{
  \alignauthor{ABM Musa \thanks{ABM Musa was a PhD candidate at University of Illinois at Chicago when this research was conducted.}}\\
\affaddr{Department of Computer Science}\\
\affaddr{University of Illinois at Chicago}\\
\email{amusa2@uic.edu} 
\alignauthor{Jakob Eriksson\thanks{This material is based upon work supported by the U.S. National Science Foundation under Grants CNS-1017877 and CNS-1149989.}}\\
\affaddr{Department of Computer Science}\\
\affaddr{University of Illinois at Chicago}\\
\email{jakob@uic.edu}
}

\newcommand{\je}[1]{\textcolor{red}{Jakob: #1}}
\newcommand{\musa}[1]{\textcolor{green}{Musa: #1}}

\newcommand{\todo}[1]{\textcolor{blue}{TODO: #1}}
\newcommand{\warn}[1]{\textcolor{red}{WARNING: #1}}
\newcommand{\note}[1]{\textcolor{cyan}{NOTE: #1}}

\maketitle

%
\begin{abstract}


  While the satellite-based Global Positioning System (GPS) is adequate for some outdoor applications, many other applications are held back by its multi-meter positioning errors and poor indoor coverage. In this paper, we study the feasibility of real-time {\it video-based} localization on resource-constrained platforms.  Before commencing a localization task, a video-based localization system downloads an offline model of a restricted target environment, such as a set of city streets, or an indoor shopping mall. The system is then able to localize the user within the model, using only video as input.

To enable such a system to run on resource-constrained embedded systems or smartphones, we 
  (a) propose techniques for efficiently building a 3D model of a surveyed path, through frame selection and efficient feature matching, (b) substantially reduce model size by multiple compression techniques, without sacrificing localization accuracy, (c) propose efficient and concurrent techniques for feature extraction and matching to enable online localization, (d) propose a method with interleaved feature matching and optical flow based tracking to reduce the feature extraction and matching time in online localization.

Based on an extensive set of both indoor and outdoor videos, manually annotated with location ground truth, we demonstrate that sub-meter accuracy, at real-time rates, is achievable on smart-phone type platforms, despite challenging video conditions. 

\end{abstract}

\section{Introduction}
\label{s:intro}

Localization technology is the key enabler of many important mobile and sensing applications today. However, the inherent limitations of current localization technology often limit the performance of current applications, and render many other infeasible. 

In an outdoor setting, the Global Positioning System (GPS) is in widespread use today.  
Consumer grade GPS receivers generally encounter an error of 5--10 meters \cite{gps_accuracy1} under ideal conditions, and over 100 meters \cite{gps_accuracy_urban1} under non-ideal conditions, such as in urban canyons where tall buildings result in an obstructed view of the sky and multi-path effects. 
Cellular \cite{cell-tower} and Wi-Fi \cite{placelab,ap-fingerprinting-kalman,metropolitan-wifi} localization can complement GPS in outdoor areas with poor or no GPS signal. 
Cellular localization often incurs errors of 100s of meters in urban areas to a few kilometers in rural areas, whereas
outdoor Wi-Fi localization is limited to urban areas due to the short range of Wi-Fi access points, and offers lower accuracy than GPS.

Indoor localization is even more challenging, as the GPS signal is typically unavailable indoors.
There has been a significant research on indoor localization using other techniques such as ultrasonic sensors \cite{cricket,active_bat}, acoustic beacons \cite{acoustic,walrus,sorroundsense}, light \cite{loc_light,sorroundsense}, and Wi-Fi \cite{radar,loc_wo_pain,loc_fine_grained,zero_cost,loc_limit,spotloc}.
Ultrasonic and acoustics sensors can offer good accuracy, 
however such methods require adding instrumentation to the space, which can be impractical. 


%

More demanding applications such as autonomous vehicles \cite{kitti-autonomous,apolloscape}, 
pedestrian navigation for the visually impaired \cite{blind-nav}, and self-guided museum tours \cite{virtualtourist}, require sub-meter localization.
Sub-meter accuracy is generally available only using expensive hardware such as survey-grade GPS receivers, high-cost IMU, and LIDAR, and is largely unavailable indoors, or in a pedestrian context. 

By contrast, video-based localization \cite{get_out_of_my_lab,wide_area_loc,loc_sfm,global_slam_loc} is applicable to both indoors and outdoors and has been shown to achieve sub-meter accuracy.
However, video-based localization can be very compute- and storage intensive.
In this work, we investigate the feasibility of real-time, sub-meter localization using a smartphone or other resource-limited platforms, through an end-to-end video localization system optimized for this target. The primary contributions of this paper are as follows.

\begin{itemize}
\item Efficient and accurate 3D model construction with video data by prioritized and filtered feature matching.
\item Compression of a 3D model by reducing both 3D points and image features to reduce storage needs.
\item Fast feature extraction by subdivision of a video frame, and parallelized, incremental feature computation.
\item Interleaved feature matching and optical flow-based feature tracking for real-time localization.
\item An end-to-end system combining multiple interdependent components involving efficient model building and online localization.
\item Evaluation of our system with meticulously annotated ground-truth data.
  
\end{itemize}

The rest of the paper is organized as follows.
We provide a system overview in \S\ref{s:overview}.
In \S\ref{s:reconstruction}, we discuss the 3D reconstruction of the environment, our contributions to achieve high-quality reconstruction, and 3D model compression.
We present the localization pipeline in \S\ref{s:online_localization}, where we discuss our contributions to make the pipeline faster to achieve near real-time operation. 
In \S\ref{s:visloc_eval_datasets} we discuss evaluation challenges and our ground truth datasets. 
In \S\ref{s:results_visloc}, we present the results of localization accuracy and other performance measures. 
Finally, we discuss the related work in \S\ref{s:related_visloc}.

\section{System Overview}
\label{s:overview}


The envisioned localization system consists of four main phases: survey, offline model creation, model retrieval, and online localization. We describe each in some detail below.

\subsection{Survey}

In the survey phase, a content creator traverses the target environment, recording a video along the way. It is not necessary to record GPS coordinates during the survey phase, although if localization within an Earth reference frame is required, this can be convenient. For large scale collection outdoors, a survey grade GPS and IMU may be used during the survey phase, to automatically tag each image with an accurate location. 

Here, the target environment could be an indoor shopping mall, the walk to the bus stop, or an individual room. How much to capture of the environment during this traversal depends on the application, but generally speaking, a more complete model of the space may results in more robust localization. It is important that any significant features, such as shop doors, elevator buttons, or crosswalks, be well captured by the video. 

\subsection{Model Creation}

In the model creation phase, the video is processed to automatically create a 3D model of the space visited. After processing, the content creator is presented with an interactive 3D model, which they annotate with any features significant to the intended navigation task. Generally speaking, the model is an independent reference frame, and localization is with respect to the model. To provide accurate model scaling, the content creator may annotate a known distance between two points within the model. If the target application needs to locate the model within the Earth reference frame, the content creator may annotate three or more points within the model with earth coordinates collected separately. Finally, if frames are tagged with accurate GPS coordinates, this may also be used. 

In more detail, we first extract keypoints and descriptors from the video frames and match descriptors among adjacent frames.
For faster matching, we build an Approximate Nearest-Neighbor (ANN) index for all descriptors of a frame and match the descriptors of adjacent frames using the index.
Second, we use the descriptor matches to reconstruct the 3D model of the environment using Structure-from-motion (SfM) \cite{bundler,bundler2,sfm-rome}.
This stage produces a 3D point cloud representing the environment along with corresponding image descriptors from video frames. 
Additionally, we construct an ANN index from all image descriptors for efficient descriptor matching during the online localization stage.

The produced model thus consists of a set of named locations, a 3D point cloud with visual features attached to each point, and an associated visual feature index.

\subsection{Model Retrieval}

Before commencing localization in a given space, a model of the space is downloaded. Model identification can be by coarse location, by name, or by any other identifier. While the incremental model download is a straightforward extension, we assume that a complete model of the space is downloaded in one step. 

\subsection{Online Localization}

Online localization continuously matches visual features in the live camera view against visual features in the 3D model, and tracks identified points in the space as the user moves. With a sufficient number of correctly identified 3D points, the camera's pose is accurately recovered. 

More specifically, we extract keypoints and descriptors from each video frame, and match these descriptors with the survey descriptors using the nearest-neighbor index.
Since we know the correspondence between 3D points and image descriptors for the offline survey, we can find the correspondence between descriptors of localization frames and 3D points. 
Using these correspondences, we compute 6 degree-of-freedom (DOF) pose/location.
Finally, we apply a Kalman filter \cite{kalman} to compute optimal poses in the presence of noise and inaccuracies.



Below, we discuss the model creation and online localization phases in more detail. 

\section{3D reconstruction of an environment}
\label{s:reconstruction}
\begin{figure*}
\centering
\includegraphics[width=0.85\textwidth]{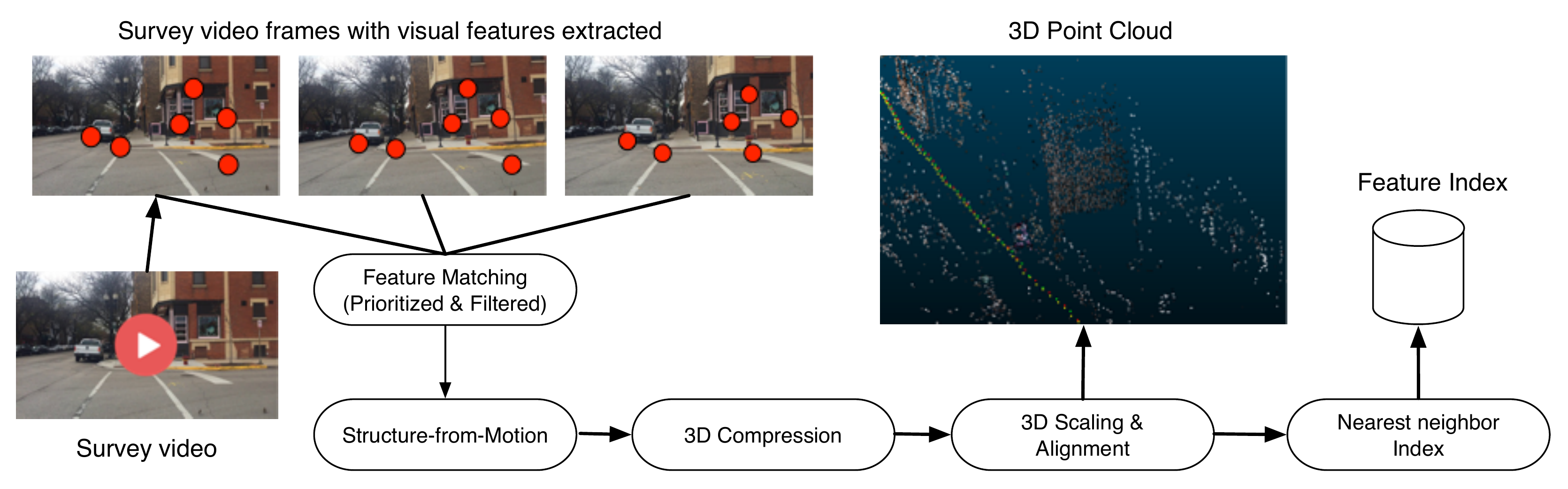}
\caption{Offline 3D reconstruction pipeline. From a video recording, a 3D model and associated index is created.}
\label{f:reconstruction_pipeline}
\end{figure*}

3D reconstruction of an environment is the most significant component of our offline survey process.
Figure \ref{f:reconstruction_pipeline} shows the stages of the 3D reconstruction pipeline.
We describe the stages of this process in more detail below.

\subsection{Structure-from-motion}
It is possible to reconstruct a 3D environment from multiple images from different viewpoints using Structure-from-motion (SfM) \cite{bundler,bundler2,sfm-rome}. SfM reconstructs the original 3D world from a sequence of images. 
This is similar to stereo-vision, where two cameras are used to infer depth, and subsequently, reconstruct an environment. 
However, unlike stereo-vision, SfM can reconstruct an environment from a single camera due to the disparity between the images resulting from camera movement. 

We reconstruct the 3D model of an environment from video recorded with a smartphone while walking or driving.
We use the open source tool Bundler \cite{bundler_url} for the reconstruction.
One important aspect of the reconstruction process is that the success and quality of the reconstruction are directly dependent on the underlying feature matching quality.
We use three heuristics to provide good matching for the reconstruction process: 

\begin{figure}
\centering
\includegraphics[width=0.45\textwidth]{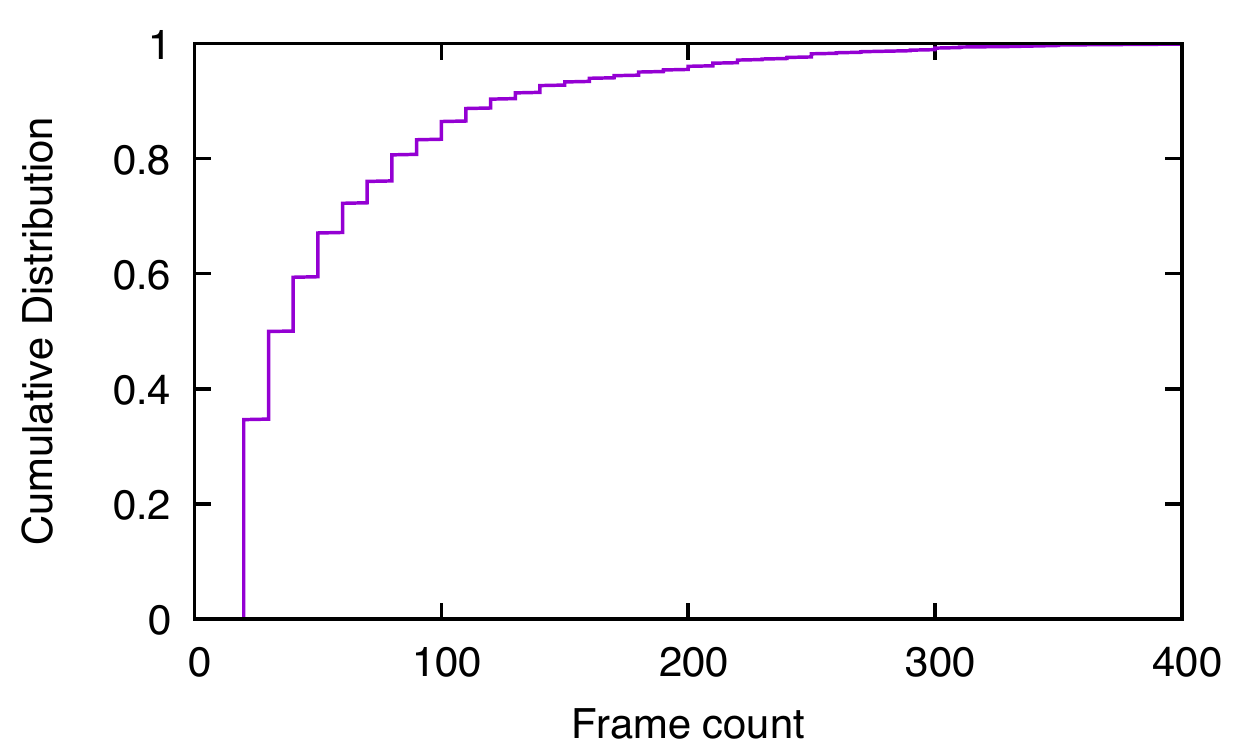}
\caption{CDF of the count of frames of each point for all points in a 3D reconstruction.}
\label{f:cdf_point_framecount}
\end{figure}

\begin{itemize}
\item We match keypoint descriptors of a frame with up to 300 adjacent frames (10 seconds of video at 30 frames/sec).
  According to our experiments, a 3D point gets reconstructed mostly from adjacent video frames.
  Figure \ref{f:cdf_point_framecount} shows an example of this.
  It shows the CDF of the count of frames that contribute to the construction of a 3D point, for all points in a reconstruction.
  Here, almost all points in the 3D reconstruction come from nearby 300 image frames.
  Additionally, a limited number of adjacent frames reduces the ambiguity that can arise from spurious matches from distant frames. 

\item Within the 300 adjacent frames, we prioritize matches for nearby frames as they are more likely to create consistent 3D points, compared to matches from more distant frames. 
Accordingly, we gradually reduce the total number of matches we keep for reconstruction from the distant frames.
Here, we use ratio test \cite{sift} (ratio=0.7) to remove ambiguous matches and order the matches based on L2 distance for a gradual reduction.

\item For successful triangulation during SfM reconstruction, it is important to have a disparity between two frames so that the triangulation converges. 
Hence, we do not use descriptor matching between adjacent frames as there is little motion or disparity between them. 
Instead, we match descriptors for regularly spaced frames. 
We experimented with various intervals such as every 5th frame, 10th frame, 20th frame, etc.
We finally used the 10 frame interval since it provided the best trade-off in terms of model size, runtime, and localization accuracy. 
\end{itemize}

%
%
\begin{figure}[ht]
\centering
\subfloat[Example location of reconstruction] {\includegraphics[width=2.3in]{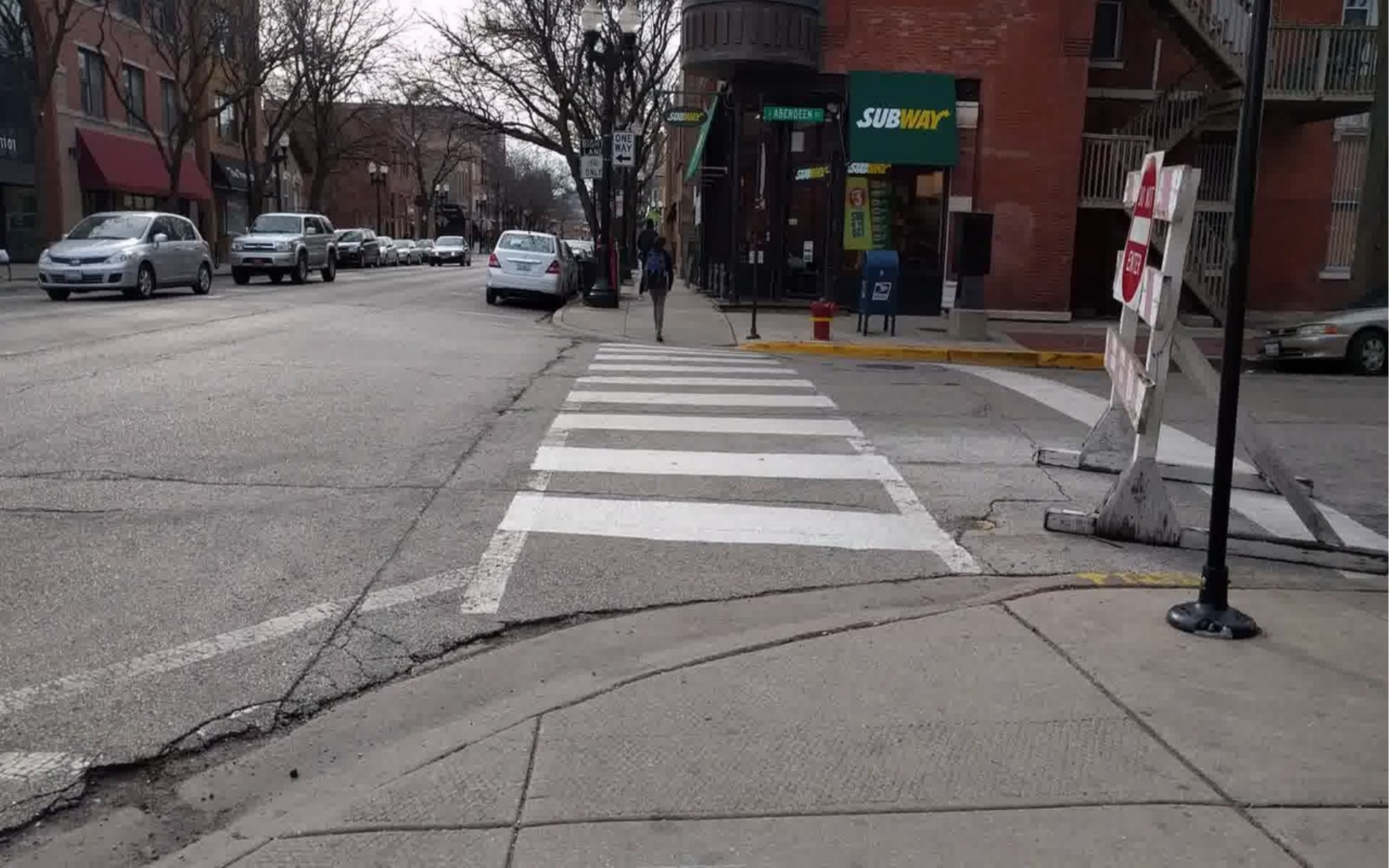}}

\subfloat[Reconstructed point cloud] {\includegraphics[width=2.3in]{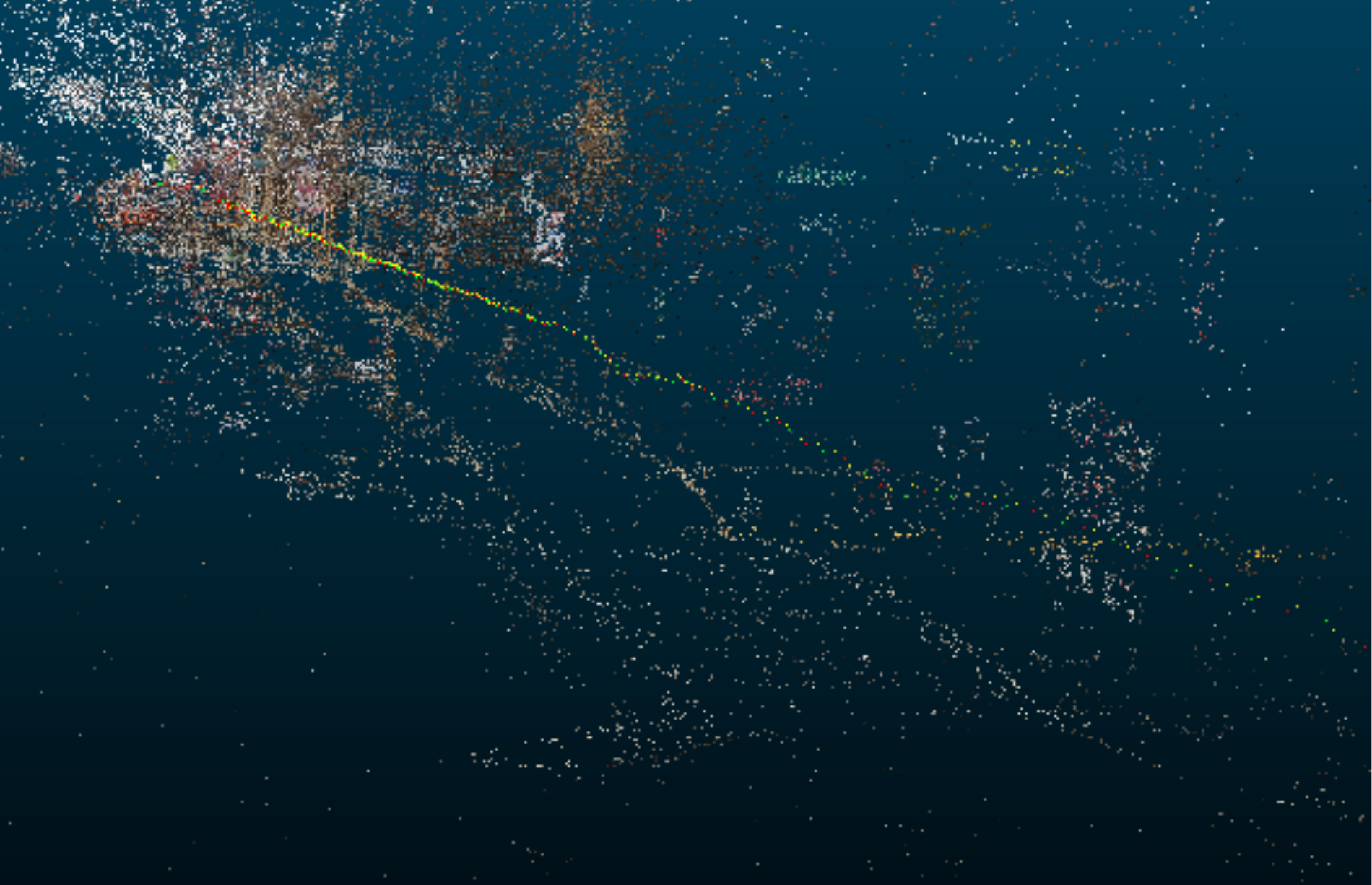}}

\subfloat[Dense reconstruction] {\includegraphics[width=2.3in]{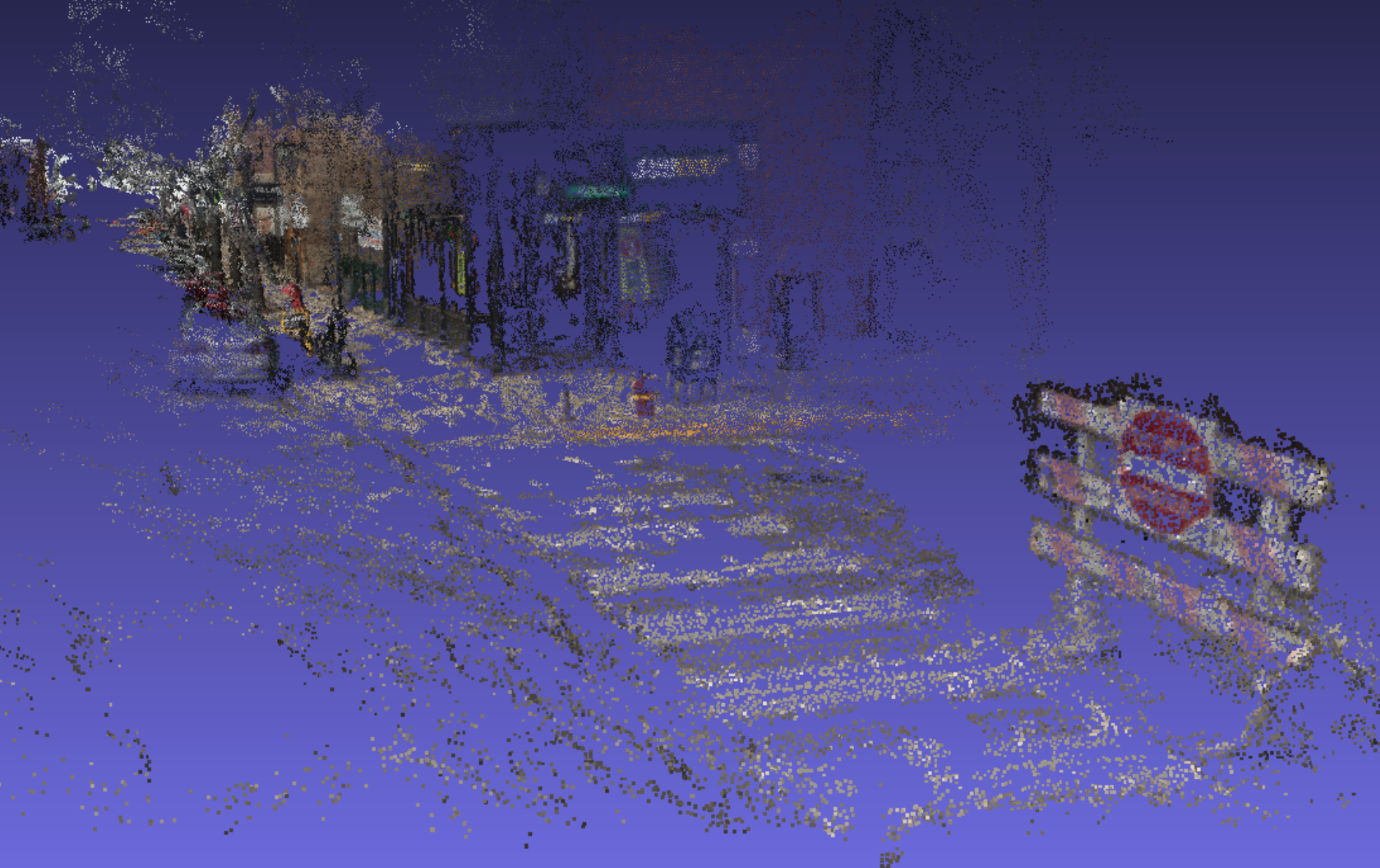}}
\caption{3D reconstruction example. The original video frames (a) are used to construct the sparse model (b) used for localization. The dense model (c) is for visualization and annotation purposes only. }
\label{f:reconstruction}
\end{figure}

Figure \ref{f:reconstruction} shows an example reconstruction. 
Here, we use a sparse point cloud reconstruction for localization purposes. 
However, we also show the dense reconstruction for visualization. 
The dense reconstruction shows that the real world is accurately represented in the 3D reconstruction.
Here, we clearly see environment details such as the stop sign, the road closure barricade, the green street sign, and the Subway logo.

\subsection{3D model compression}
\label{s:3d_model_compression}
3D model compression reduces both storage and computational needs by removing redundant data from the model. 
We use two techniques for 3D model compression: 1) reduction of 3D points of the model, 2) reduction of descriptors by averaging all descriptors corresponding to a 3D point.

\begin{figure}[ht]
  \centering
  \begin{minipage}[b]{0.45\textwidth}
    \includegraphics[width=\textwidth]{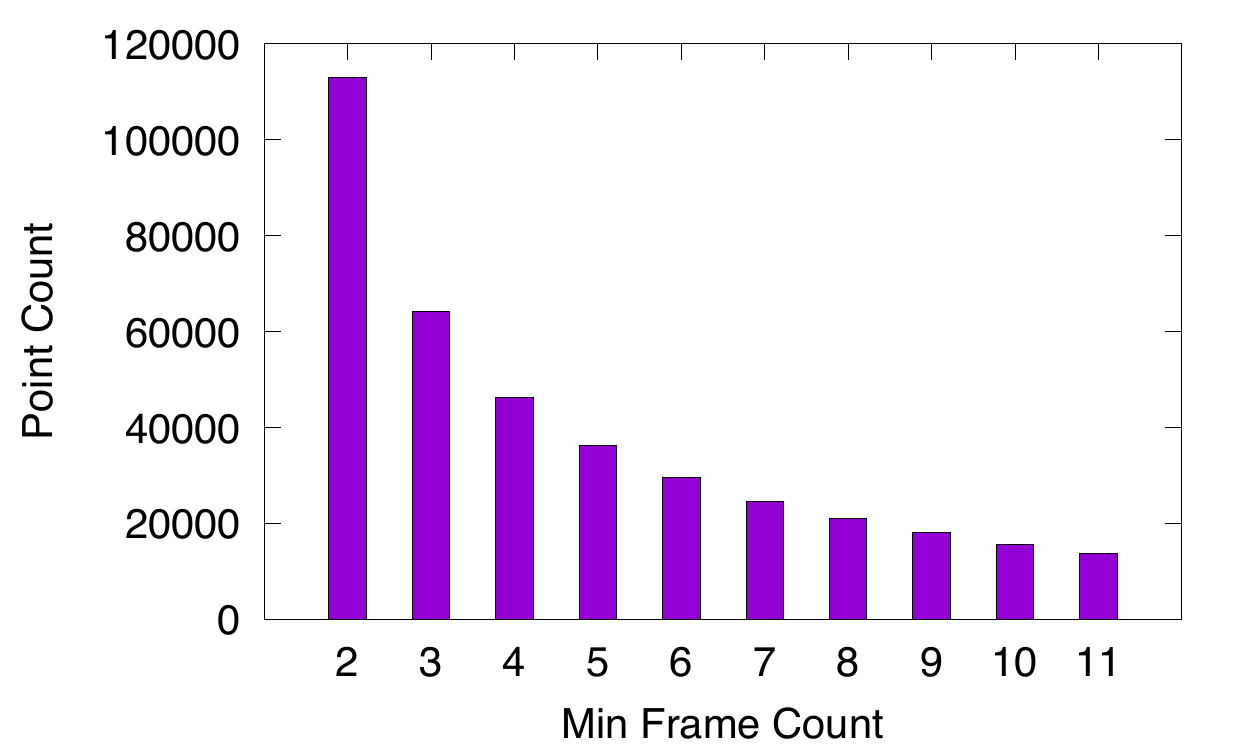}
    \caption{3D point count in a model for minimum frame correspondences.}
    \label{f:3d_point_reduction}
  \end{minipage}
  \hfill
  \begin{minipage}[b]{0.45\textwidth}
    \includegraphics[width=\textwidth]{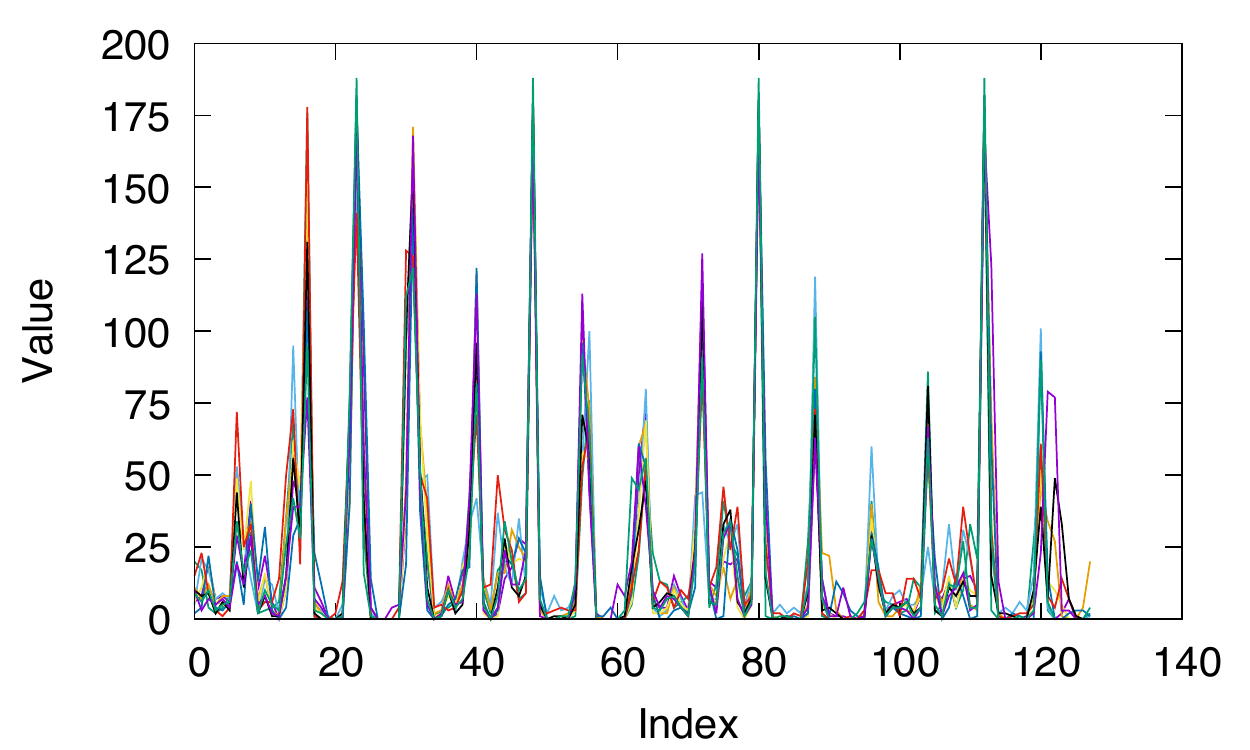}
    \caption{SIFT descriptor vector values for matching descriptors corresponding to a 3D point.}
    \label{f:feature_similarity}
  \end{minipage}
\end{figure}

The 3D reconstructed model obtained using Structure-from-Motion contains a large number of 3D points and their corresponding descriptors, consuming significant memory.
For example, one of our 3D reconstruction from 700 images produced 113,000 3D points, and 630,000 distinct, 128-dimensional descriptors.
Many of these 3D points only correspond to a small number of frames.
Figure \ref{f:3d_point_reduction} shows the 3D point count for various minimum corresponding frame count.
Here, approximately 46\% of the total points appeared in only 2--3 frames. 
Accordingly, we can substantially compress the 3D model by removing 3D points that correspond to a small number of frames.
After all, we only need a small number of correct correspondences from the 3D model during the localization, and points that are rarely seen in the survey phase are unlikely to be commonly observed during online localization.

Moreover, the descriptors corresponding to a 3D point are by definition very similar. 
Figure \ref{f:feature_similarity} shows an example of this. 
Here, we show the component values of the SIFT descriptor vectors for matching descriptors corresponding to a 3D point.
Since all these matching descriptors are very similar, we replace them all by the mean descriptor of the point, resulting in additional compression of the 3D model.

Table \ref{tab:model_storage} shows the storage requirement for a 3D model constructed from video of a 300-meter long urban street.
Here, the storage requirement after two-stage compression is 7.5MB compared to the uncompressed storage of 91MB, or a 12$\times$ size reduction. 
As we show in (see \S\ref{s:compression_effect}), this type of 3D model compression has a negligible effect on localization performance.

\begin{table}
\centering
\begin{tabular}{lll}
3D points & Descriptors & Storage \\
\hline
All & All & 91 MB \\ 
At least 10 frames  &  All & 41 MB  \\ 
At least 10 frames  &  Mean & 7.5 MB  \\ 
\hline
\end{tabular}
\caption{Storage requirement for a 3D model representing a 300 meter long urban street.}
\label{tab:model_storage}
\end{table}

\subsection{3D scaling and alignment}
\label{s:alignment}
The 3D reconstruction obtained using SfM typically has an arbitrary scale and orientation. 
All 3D points are consistent relative to each other within the model, but absent scaling and alignment, they do not represent real-world dimensions.
For the most accurate alignment with the earth coordinate system, the content creator can specify the precise coordinates of three points within the model. 

If accuracy of alignment is less important, an outdoor deployment can leverage GPS coordinates from the survey video. 
While the instantaneous accuracy of a consumer grade GPS is low, averaging over hundreds of GPS samples spanning a longer recording will eliminate much of the error.
If needed, the survey can be performed with a survey grade GPS.  

For our evaluation purposes, we use the ground-truth scale and orientation information recorded during the survey. 
We describe our approach to ground-truth data collection in \S\ref{s:visloc_eval_datasets}.

\subsection{Index for efficient matching}
Each 3D point in the reconstructed model has a corresponding set of matched image descriptors.
During localization, we match image descriptors of a localization frame with the descriptors of survey frames to find the 3D point correspondence.
Matching a localization frame with a subset of reference frames in a one-to-one fashion is prohibitively expensive. 
Hence, we construct an approximate nearest-neighbor (ANN) index using the FLANN \cite{flann1,flann2,flann3} library with all descriptors of the 3D model.
Approximate matching with the ANN index provides adequate accuracy for localization.

eifjccinklcgbcgvcuuekchfhetnunvljivuriintiug
\section{Online localization}
\label{s:online_localization}
Below, we discuss the design of the online localization pipeline, with the goal of real-time operation.

\begin{figure*}
\centering
\includegraphics[width=6.0in]{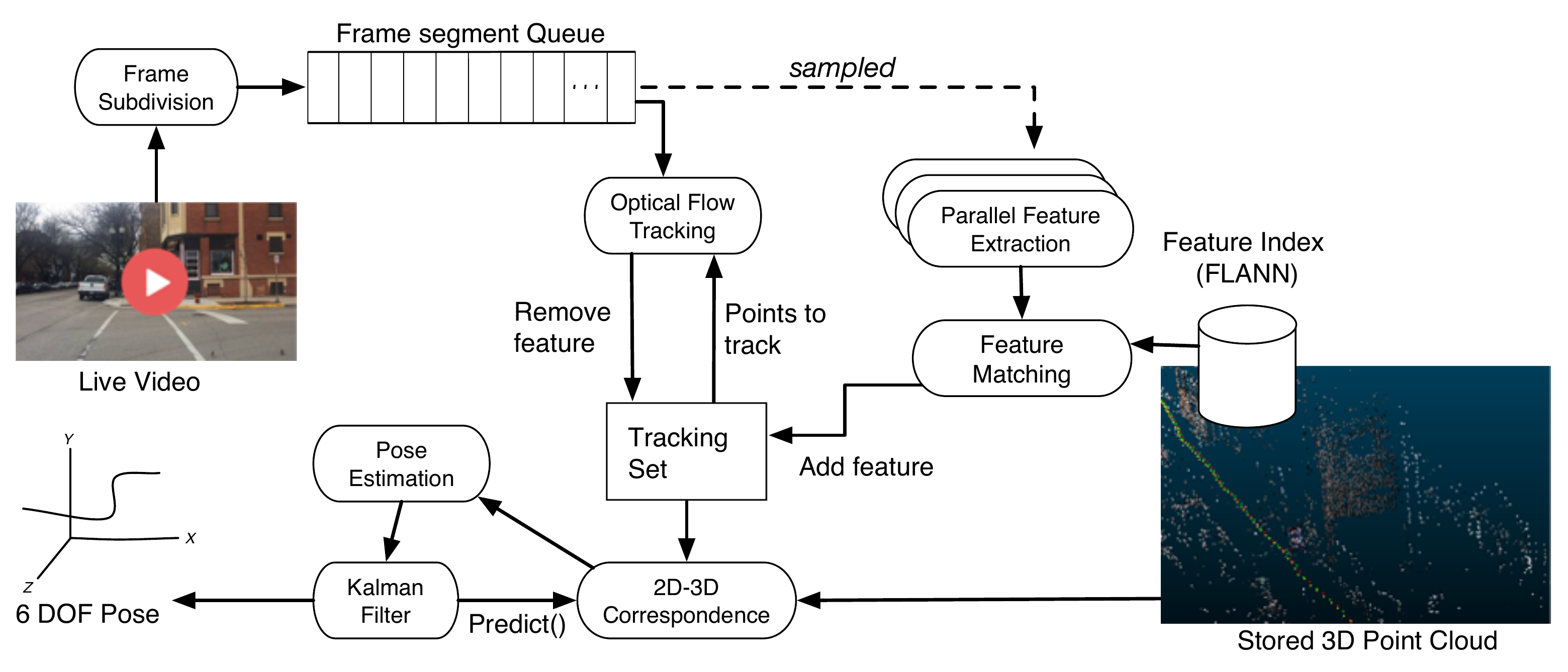}
\caption{Online localization pipeline. Feature extraction and matching on sampled frames populates a tracking set, which is maintained by optical flow tracking. The points in the tracking set all contribute to the pose computation. }
\label{f:localization_pipeline}
\end{figure*}

Figure \ref{f:localization_pipeline} shows our localization pipeline in detail. Throughout, the focus is on minimizing latency, through pipelining, parallelization, and sampling.
To provide a high rate of localization, we separate localization into parallel {\it matching} and {\it tracking} processes. 

 {\it matching}  identifies key points in the image, and matches these against the 3D model. This is the key operation in the localization pipeline, but it is computationally both highly variable and quite demanding, consuming 1--2 seconds per frame on a server core. To support a typical video frame rate, we parallelize the matching process, and adaptively sample incoming frames to match available computational resources. Matching thus slowly, but continuously adds key points to the tracking set.

 {\it tracking}  quickly tracks key points already in the tracking set, using optical flow. Optical flow is fast, but not very robust. Typically, we are able to successfully track a point for 50--100 frames before tracking fails, and the point is removed from the tracking set. 

Given a tracking set of 2D points, we compute their correspondence to the 3D points in the point cloud, then use a RANSAC technique to estimate the final 6-DOF pose. 
Finally, we filter the stream of poses using a Kalman filter to remove noise and other aberrations.
The Kalman filter output is also used to inform the correspondence computation.
We describe these stages in more detail below.

\subsection{Keypoint and descriptor computation}
SIFT keypoint and descriptor computation generally takes 1 to 2 seconds for a 1080x720p video frame on a server CPU, and approximately 10$\times$ longer on a smartphone core. 
We use two techniques to amortize this computation time. 
First, we subdivide a video frame into smaller segments and compute SIFT keypoints and descriptors for some of these segments in parallel.
In our experiments, we use 8 cores for parallel computation since many recent embedded processors and smartphones have 8 cores.
Second, the frame arrival rate from a camera is still far greater than the SIFT computation rate by parallel threads. 
However, we do not need to compute keypoints and descriptors for all segments as they arrive for successful localization. 
Accordingly, we compute the SIFT keypoints and descriptors on a sampled subset of frames, and interleave this with optical flow tracking (described in \S\ref{s:tracking}).

\subsection{Descriptor matching}
\label{s:pipeline_matching}
SIFT descriptor matching is one of the most expensive stages of the localization pipeline. 
To find a descriptor's most similar feature in the point cloud (i.e., \textit{exact} nearest neighbor), the brute-force method is to compute the L1 norm or the L2 norm between the descriptor and all other descriptors and picking the descriptor with the minimum distance. 
While brute-force matching guarantees the best accuracy, it is not practical due to excessive computation time.

For fast matching, we create an Approximate Nearest-Neighbor (ANN) index for all survey video frames within an area (e.g., 100 meters) in the offline modeling phase.
Using the index, we match the localization frame with all survey frames in a single step.
Although this method can result in some inaccuracy, it is sufficient for localization as we can filter out spurious matches in later stages.
However, even this matching scheme requires over 1 second per frame on a server core. Thus, as in the keypoint and descriptor computation, descriptor matching is performed on sampled frames, and in the background.


\subsection{Keypoint tracking}
\label{s:tracking}

\begin{figure}
\centering
\includegraphics[width=0.45\textwidth]{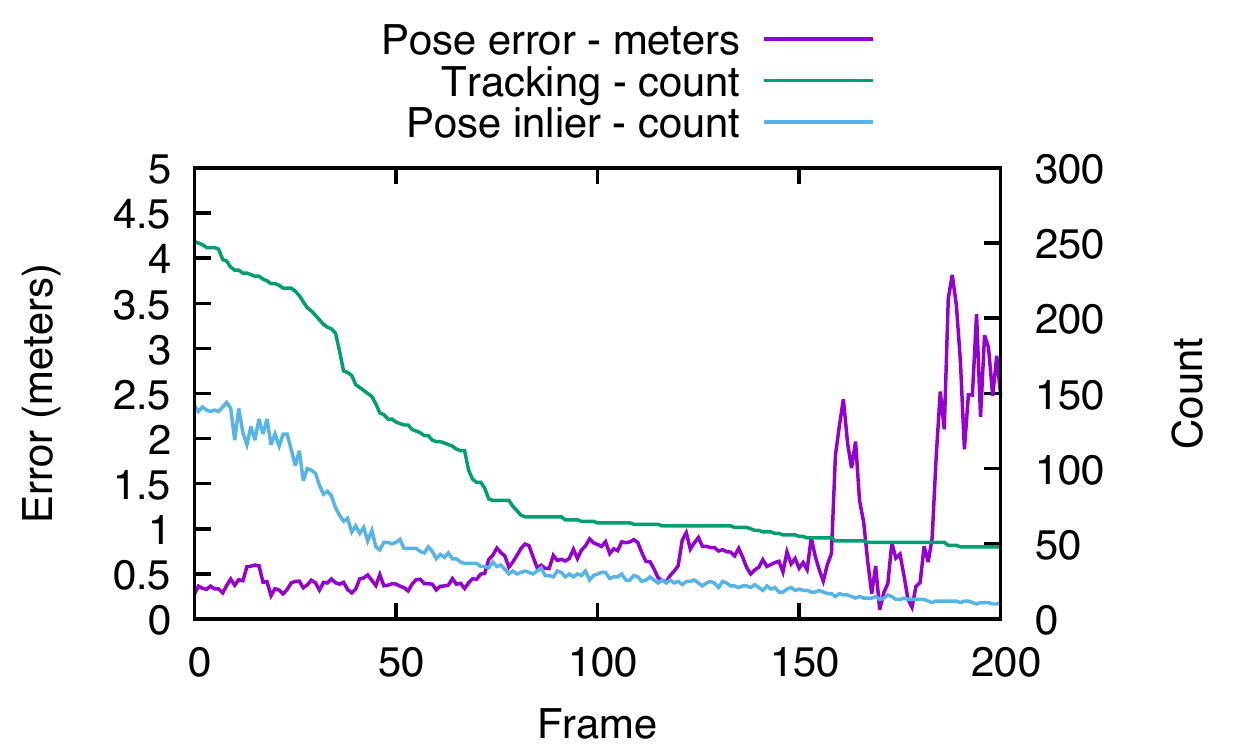}
\caption{Count of tracked keypoints (right y-axis) and count of RANSAC pose estimator inliers (right y-axis) along with pose error (left y-axis) for consecutive frames.}
\label{f:tracking_duration}
\end{figure}

\begin{figure}
\centering
\includegraphics[width=0.45\textwidth]{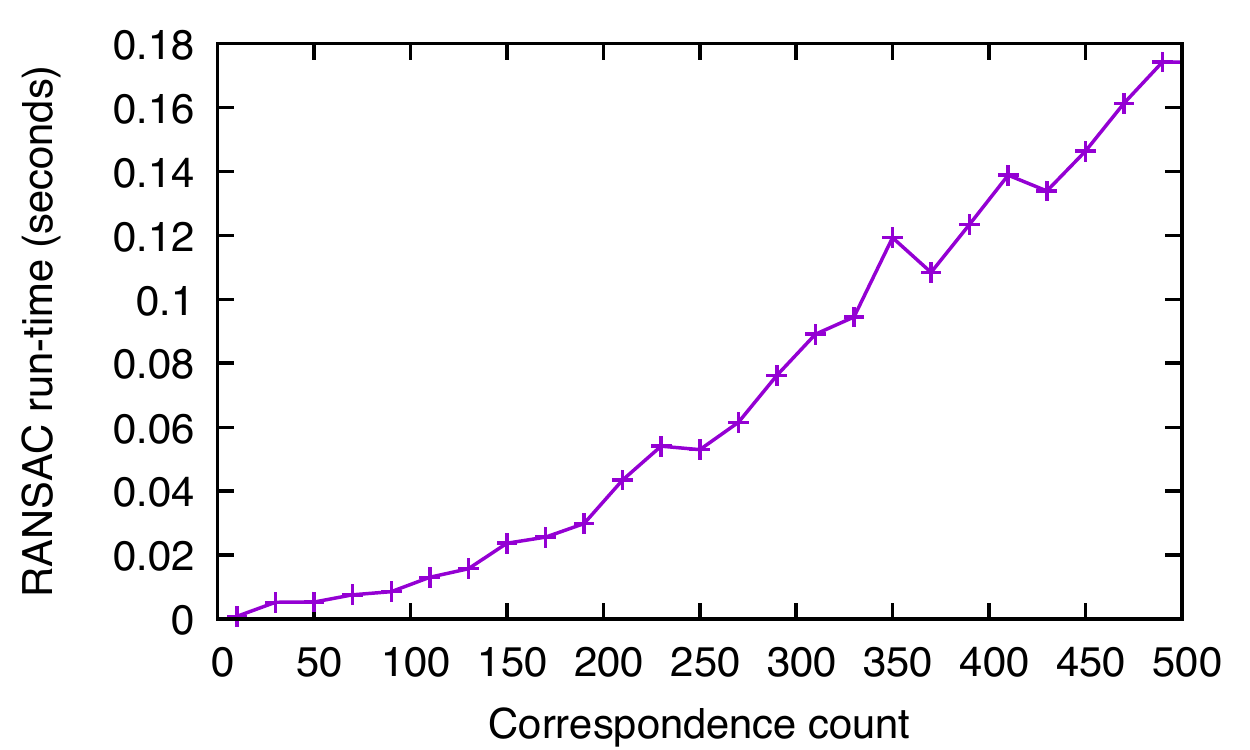}
\caption{RANSAC computation time for correspondence count.}
\label{f:ransac_time}
\end{figure}


Given the coordinates of some points in one video frame, optical flow \cite{optical_flow} computes the new coordinates of those points in a successive frame caused by the movement of the object or camera.
Optical flow is fast because it only needs to search the adjacent pixels of a point for the new location.

Tracking with optical flow works well primarily in constant lighting conditions and with a small changes to the viewpoint.
By contrast, descriptor matching is robust to varying lighting conditions. 
After localization of a video frame, we are able to use optical flow tracking for subsequent frames as the lighting condition is likely to remain the same for some time.
Eventually, however, the optical flow tracking is bound to fail for some points because of the change in viewpoint. 
To compensate for this, we continuously replenish the working set with new matched points. 

Figure \ref{f:tracking_duration} shows the frame count/duration of successfully tracked keypoints using optical flow.
Here, the left y-axis shows error in pose estimation with only optical flow tracking for consecutive frames. 
The right y-axis shows the count of successfully tracked keypoints and also the count of pose estimation inliers for consecutive frames. 
Over time, the number of successfully tracked points (and inliers) drops, and as a result, localization accuracy using only optical flow collapses after approximately 150 frames (or 5 seconds, at 30 Hz frame rate). 


\subsection{2D-3D Correspondence and pose estimation}
\label{s:pose_2d_3d}
We find the correspondence between localization frame keypoints and the 3D model points by combining the matching of localization frame descriptors with the surveyed model's descriptors, and the mapping between survey descriptors and 3D model points.
The correspondence count between a single frame and the 3D model can be large, as a 1280x720 video frame can have several thousand SIFT descriptors and a majority of them may match with the model descriptors if we are in a well-surveyed location. 

Perspective-n-point algorithms \cite{p3p,epnp} may be used to compute a 6-DOF pose/location.
These algorithms can compute pose with just 3 accurate correspondences \cite{p3p} or a few more \cite{epnp}.
However, some of our correspondences are likely to be spurious, due to errors in matching and tracking. 
To filter out such spurious correspondences, we use a RANSAC \cite{ransac,ransac_eval,optimal_random_ransac} pose estimator, which iteratively finds the correct pose.

\begin{figure}
\centering
\includegraphics[width=0.45\textwidth]{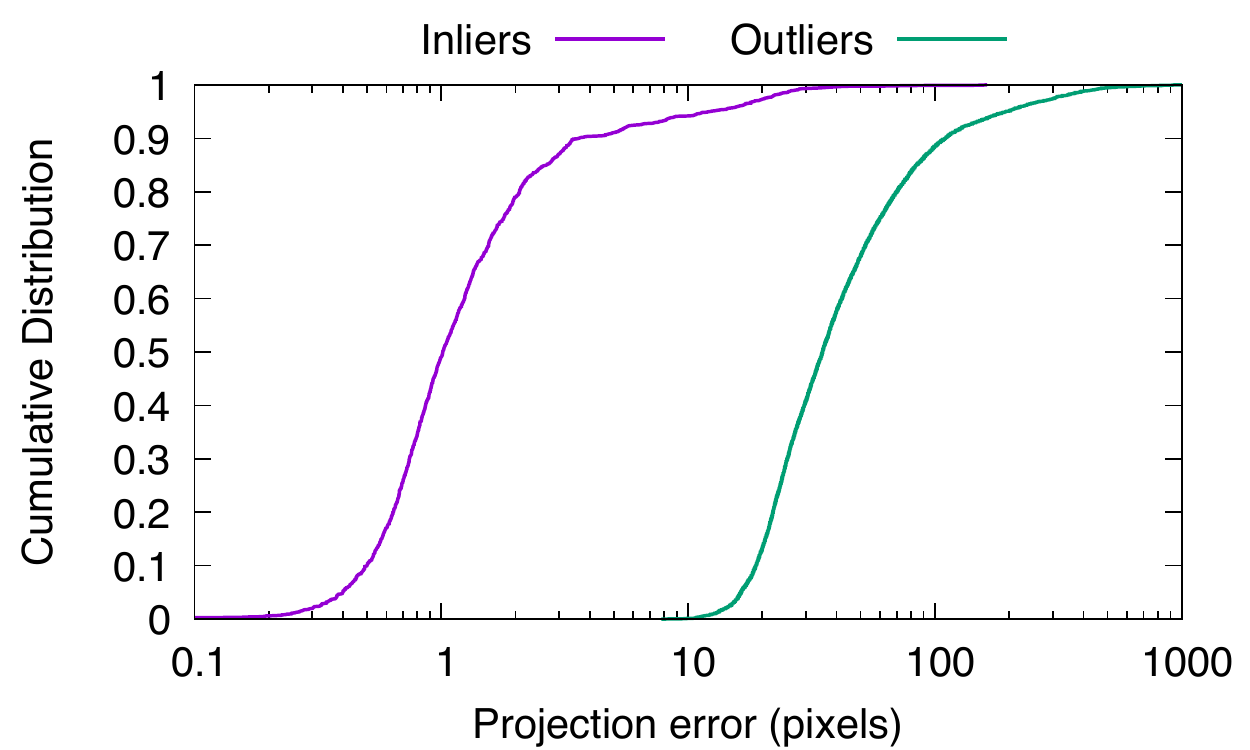}
\caption{CDF of projection error.}
\label{f:proj_error}
\end{figure}

The RANSAC pose estimator works with hundreds of correspondences and will successfully remove false correspondences (outliers).
However, working with a smaller set of correspondences is beneficial since RANSAC pose estimation converges faster with a smaller number of correspondences.
Thus, we prioritize the correspondences by quality during descriptor matching in order to take the top $k$ correspondences for RANSAC pose estimation.
To approximate the quality, we use two parameters: projection error and descriptor distance.
At first, we find the projection error (the difference between the projection of 3D points onto the image plane and the corresponding keypoints) using the Kalman filter's prediction of the next pose.
The projection error can be used to filter out erroneous matches.
For example, Figure \ref{f:proj_error} shows the CDF of projection error of both inliers and outliers for an image frame.
Here, 95\% of the inliers have projection error below 10 pixels and most of the outliers have projection error above 10 pixels.
Accordingly, we discard the correspondences that have a projection error above 10 pixels. 

Next, we order the remaining correspondences using the matching distance so that we can pick the top $k$ for pose estimation.
We use $k=100$ in this paper. Figure \ref{f:ransac_time} shows the mean computation time of RANSAC pose estimator with varying correspondence count, which shows that the time required by RANSAC reduces on average as the total number of correspondences is reduced.

\subsection{Pose correction using Kalman filter}
\label{s:pipeline_kalman}
RANSAC pose estimation can be invalid due to a number of reasons: incorrect descriptor matching, bad optical flow tracking, erroneous RANSAC output, etc.
However, the poses of successive video frames are expected to be correlated since we make a small movement between two consecutive frames while walking or driving.
To take advantage of this, we apply a linear Kalman filter \cite{kalman, kalman-tutorial} to the estimated poses of the video frames. 

We model the state vector \cite{kalman_model} with 6-DOF position $(x,y,z,roll,$ $pitch,yaw)$ and their first and second derivatives (velocity and acceleration respectively).
Thus, the state vector is ($\mathbf{x}_k$) as following: $\mathbf{x} = [x,y,z,\dot{x},\dot{y},\dot{z},$ $\ddot{x},\ddot{y},\ddot{z},\phi,\theta,\psi,\dot{\phi},\dot{\theta},\dot{\psi},\ddot{\phi},\ddot{\theta},\ddot{\psi}]^T$. Here, $x,y,z$ are position components and $\phi,\theta,\psi$ are roll, pitch, yaw.
We use a conventional physics-based process model ($\mathbf{F}$) relating the state vector between consecutive time steps (omitted for brevity).  

Our measurement model is $\mathbf{z}_k = [x,y,z,\phi,\theta,\psi]$, which we obtain from the pose estimation step as described in \S\ref{s:pose_2d_3d}.
%
%
%
Finally, we use error-gating to improve robustness to large errors in matching, tracking, or pose estimation.
With the error gating, if any pose component is not within four std-dev of the current Kalman estimated pose, we discard the measurement. 
Note that having a two or three std-dev bound is aggressive for Kalman filter for noisy data. 
In practical systems, it is more common to use four or even five std-dev bound \cite{kalman-tutorial}.

\begin{figure}
\centering
\includegraphics[width=0.45\textwidth]{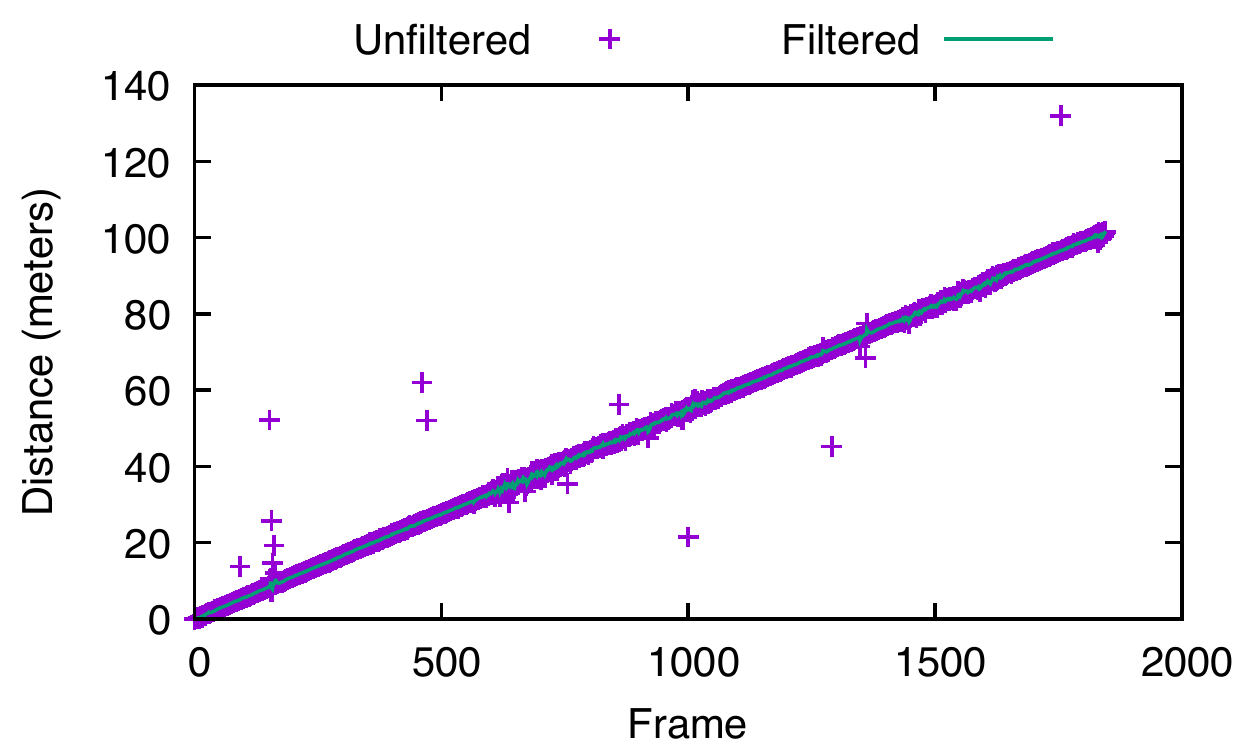}
\caption{Example of Kalman filter discarding noisy locations/poses.}
\label{f:kalman}
\end{figure}

Figure \ref{f:kalman} shows an example of the Kalman filter correcting the noisy locations.
It shows the successive distance from the starting position for 2000 frames after location/pose estimation.
The locations obtained after feature matching, optical-flow tracking, and RANSAC pose-estimation can be noisy and spurious for some frames.
The Kalman filter smooths out and corrects such noisy poses.

\section{Evaluation Challenges and Datasets}
\label{s:visloc_eval_datasets}
We discuss the evaluation challenges and two ground truth datasets that we collected overcoming these challenges in this section.

\subsection{Evaluation challenges}
\label{s:eval_challenges}
Collecting ground-truth data for sub-meter localization at a large scale is challenging. 
The typical method of collecting location data is to use a GPS device to collect latitude-longitude for geo-tagging.
However, consumer-grade GPS devices typically encounter 10s of meters of errors, and even 100s of meters in challenging environments (e.g. urban canyons, tree shades, etc.), and the challenging environments are most appropriate for video-based localization. 
Therefore it is not possible to use generic GPS receivers for ground-truth data for evaluating a system having sub-meter accuracy. 
One option to achieve high accuracy is to use Real-Time Kinematic (RTK) GPS receiver along with IMU for centimeter accurate geo-location while moving.
However, these systems are highly expensive and they do not work indoors.
Hence, in this work, we use manually annotated ground truth datasets in challenging outdoor environments and indoors for evaluation.
We used two methods to collect ground truth data.
We describe them below.

\subsection{Ground truth with a measuring wheel}
\label{s:ground_truth_wheel}
To collect ground truth data with high accuracy, we walked along a path with a measuring wheel and put chalk marks every 5 feet for a total length of 1000 feet outdoors at an urban street and 430 feet indoors in UIC campus cafe. 
We then walked the path and recorded a video for 3D reconstruction. 
Later we walked the path two more times and took pictures at every mark to compare their location for measuring the accuracy of our methods.
This dataset provides intermittent but highly accurate ground truth.

\begin{figure}
  \centering
  \begin{minipage}[b]{0.45\textwidth}
    \includegraphics[width=\textwidth]{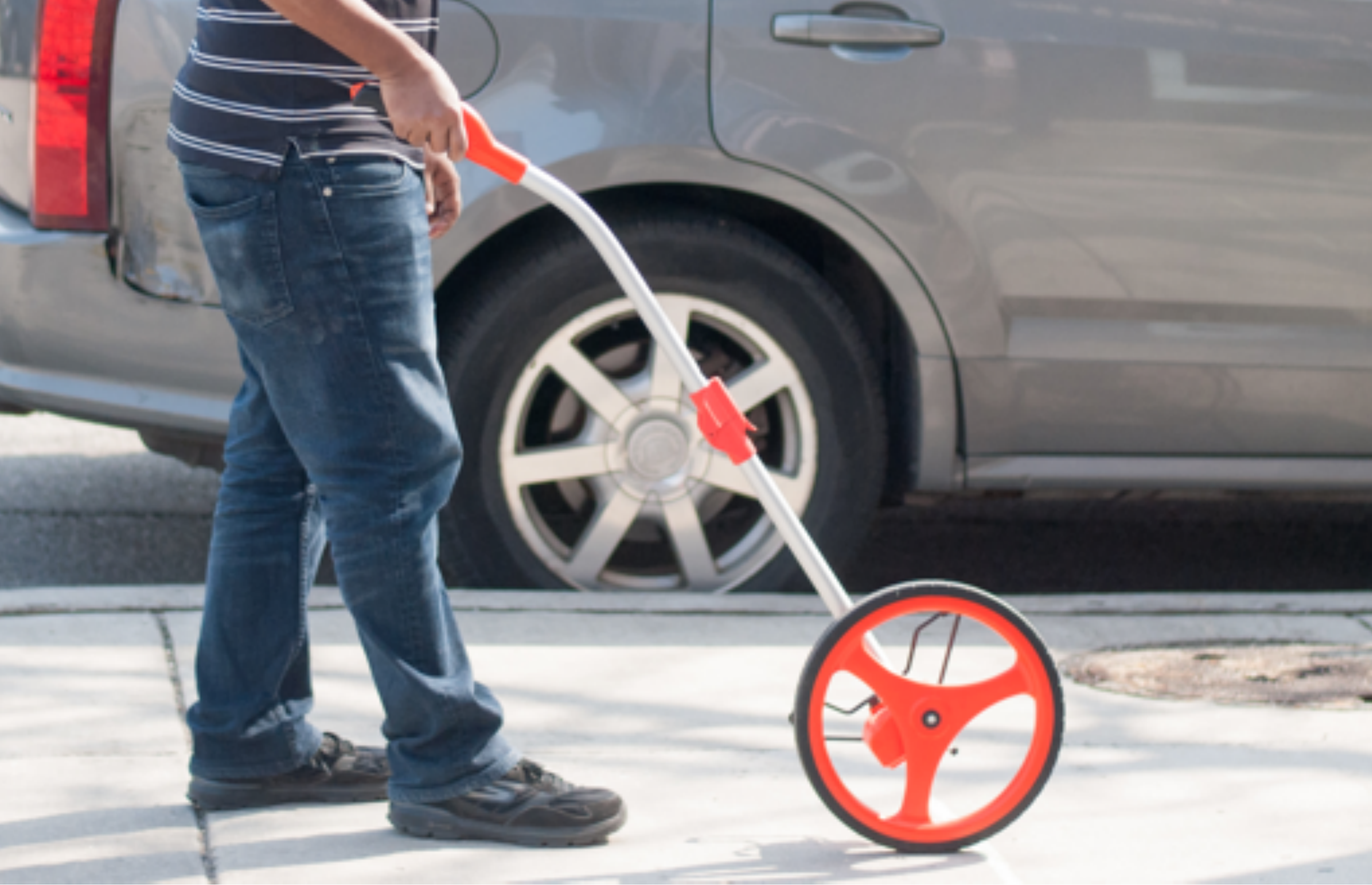}
    \caption{Ground truth data collection with measuring wheel.}
    \label{f:ground_truth_wheel}
  \end{minipage}
  \hfill
  \begin{minipage}[b]{0.45\textwidth}
    \includegraphics[width=\textwidth]{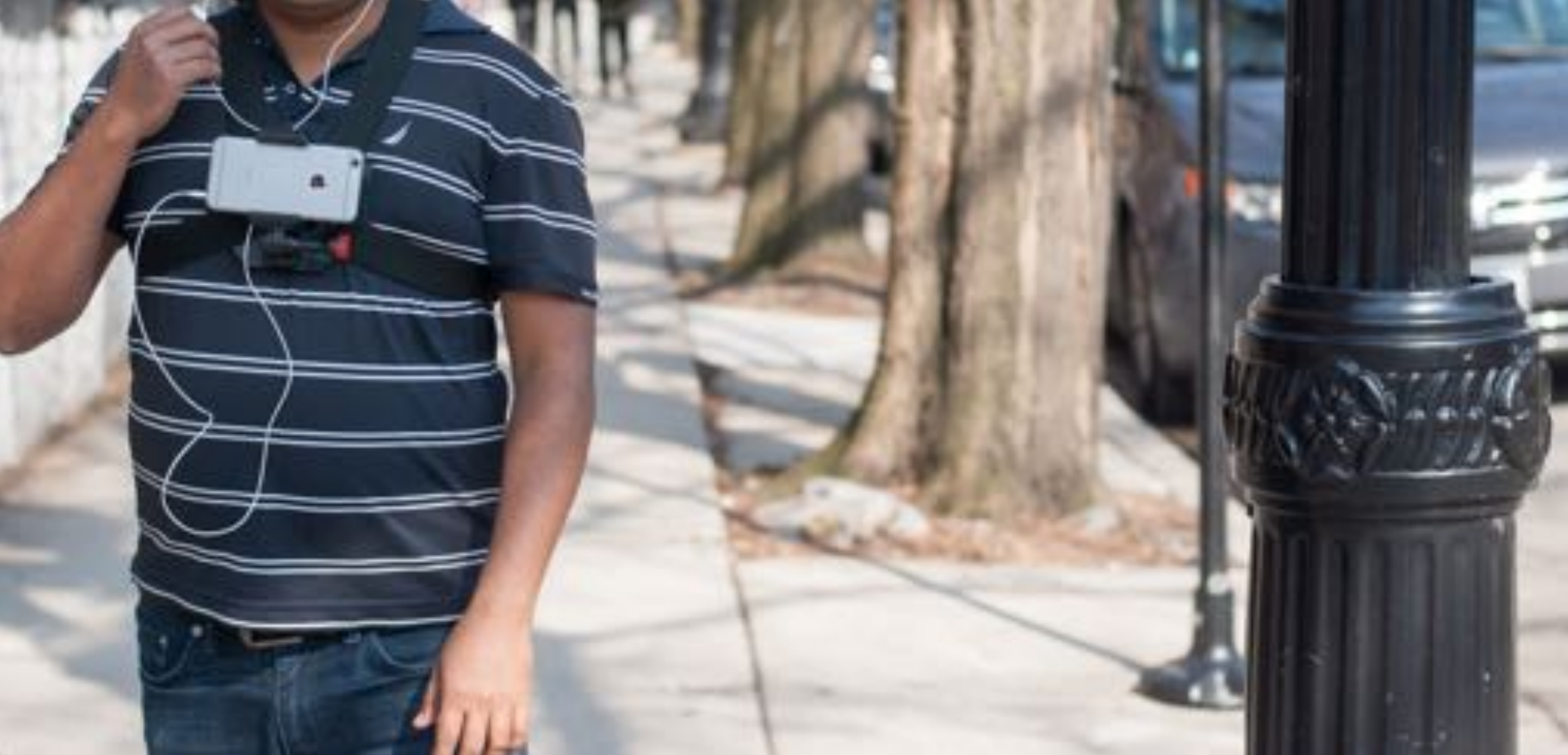}
    \caption{Ground truth data collection with chest harness.}
    \label{f:ground_truth_chest}
  \end{minipage}
\end{figure}

\begin{figure*}
\centering
\subfloat[Urban street for evaluation.] {\includegraphics[width=3.1in]{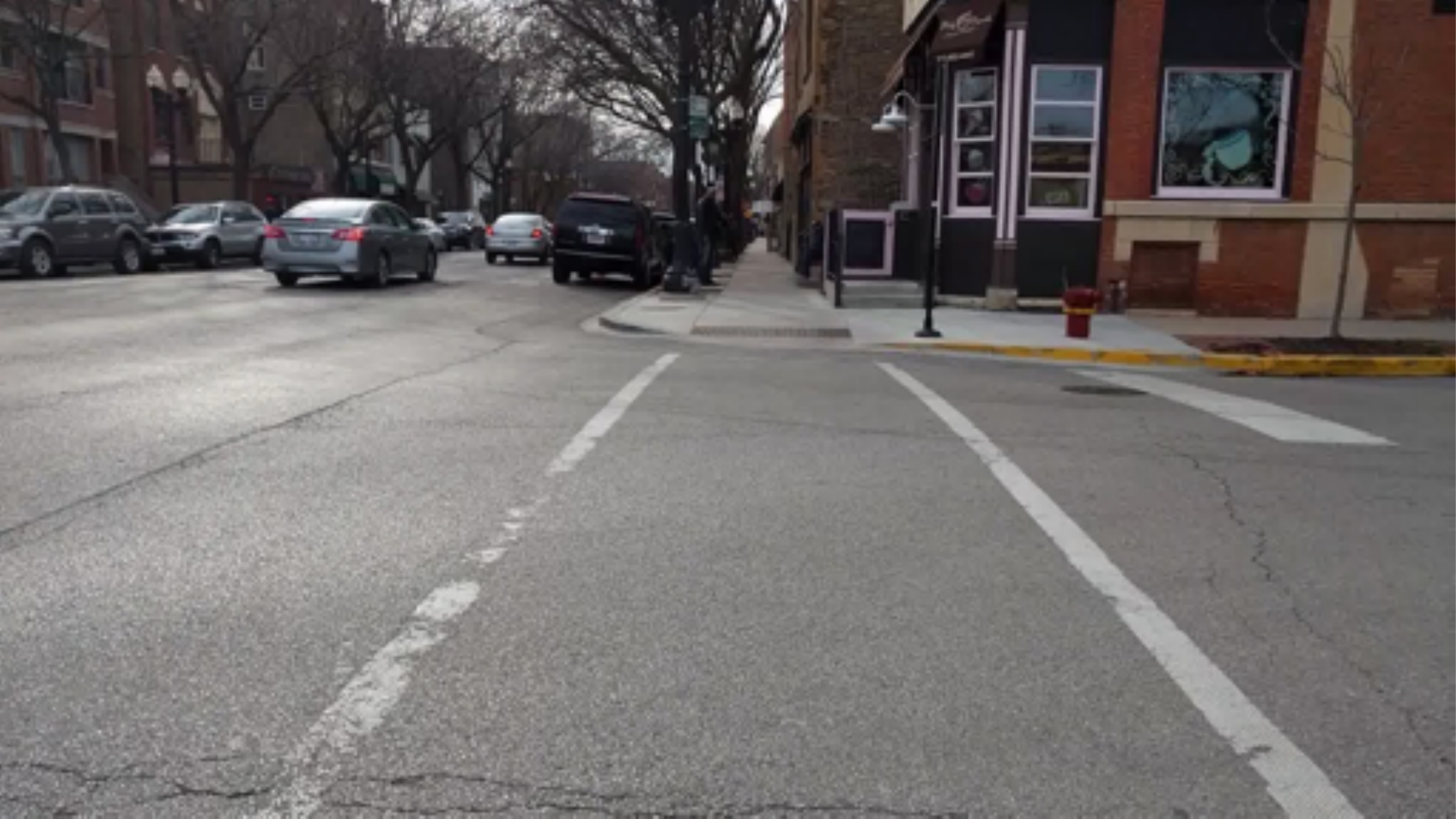}}
\vspace{0.2in}
\subfloat[Indoor cafe for evaluation.] {\includegraphics[width=3.1in]{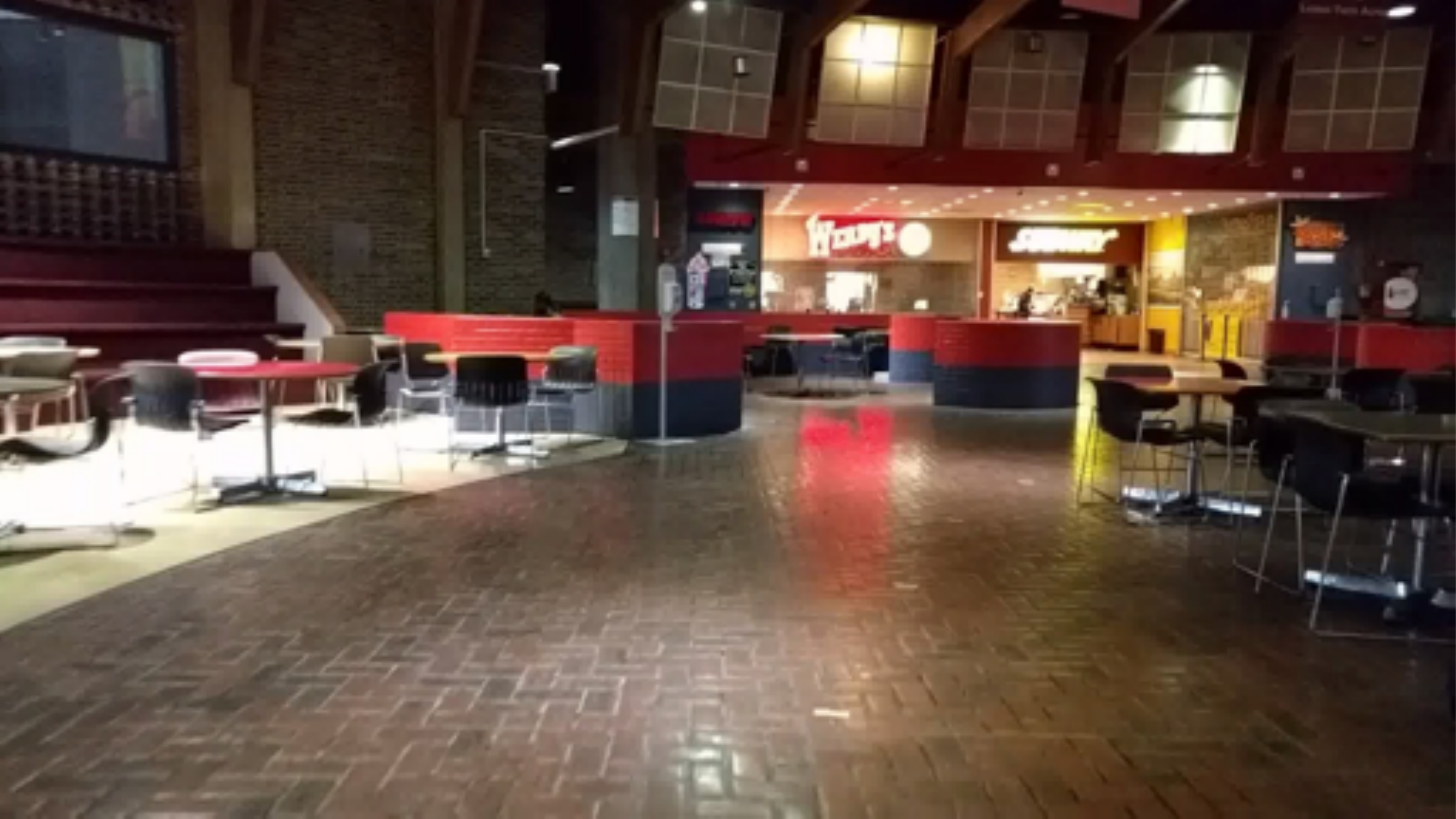}}

\caption{Evaluation Scenarios.}
\label{f:eval_scenarios}
\end{figure*}

We used a Nexus 5x smartphone for collecting these data sets. 
Figure \ref{f:ground_truth_wheel} shows the measuring wheel that we use to mark a path.
Figure \ref{f:eval_scenarios} shows the outdoor urban street and the indoor cafe at which we evaluated our system.


\subsection{Ground truth with landmark tagging}
\label{s:ground_truth_click}
The data collected with the measuring wheel offers ground truth for every 5 feet only.
However, we need to evaluate our system for continuous video frames too.
Hence, we collected another ground-truth dataset by recording video several times with an iPhone 6 at an outdoor street for measuring localization accuracy of continuous video frames.
Here, we annotated every light pole in the street by clicking the headphone button to record a time-stamp in the video as we pass by light-poles. 
The light poles are approximately 20 meters apart.
For all frames between the clicks/light-poles, we interpolate the location based on a best effort attempt of walking at a constant pace.
Since the distance between two adjacent light poles is low, this interpolation likely does not introduce a significant error.
Figure \ref{f:ground_truth_chest} is an example of this approach where we mount the smartphone using a chest harness. 





\section{Results}
\label{s:results_visloc}
We present the performance of our localization system for both intermittent and continuous ground truth below.
Additionally, we describe the effect of model compression on localization accuracy and smartphone computation time.

\subsection{Localization accuracy for intermittent ground truth}

\begin{figure*}
  \centering
  \begin{minipage}[b]{0.32\textwidth}
    \includegraphics[width=\textwidth]{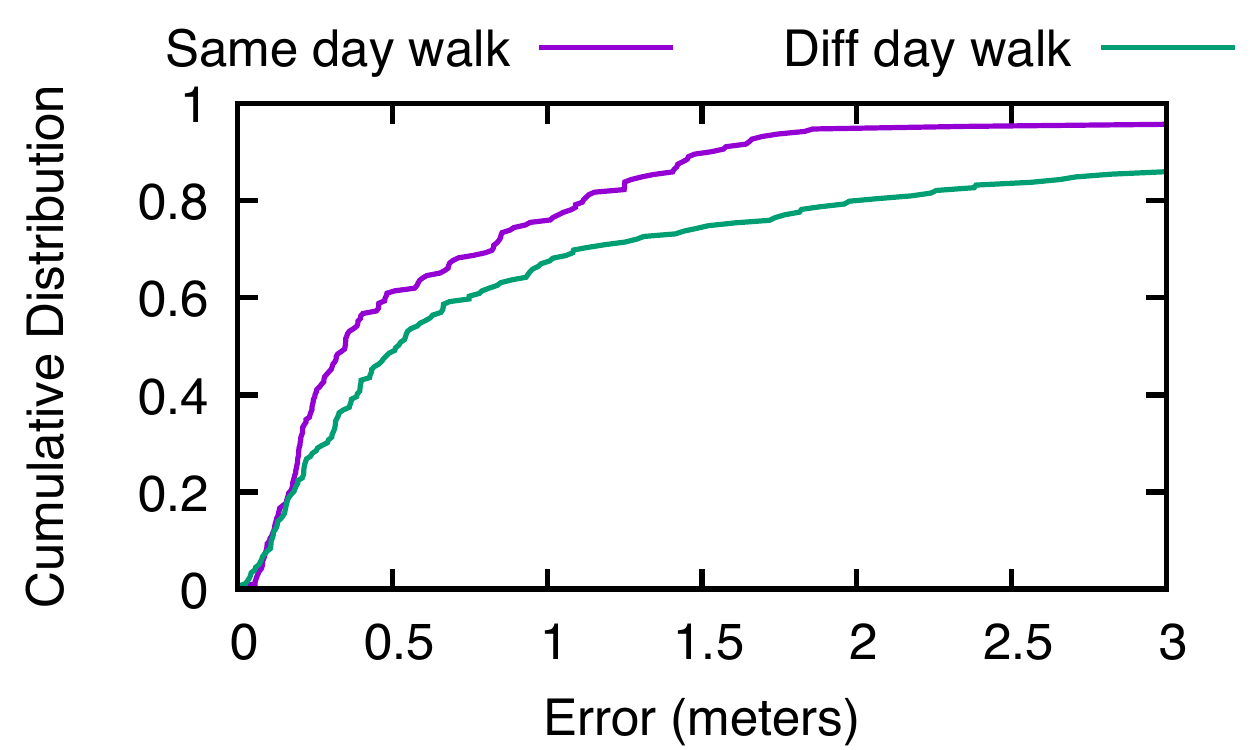}
    \caption{CDF of localization error in outdoor (intermittent).}
    \label{f:accuracy_wheel_outdoor}
  \end{minipage}
  \hfill
  \begin{minipage}[b]{0.32\textwidth}
    \includegraphics[width=\textwidth]{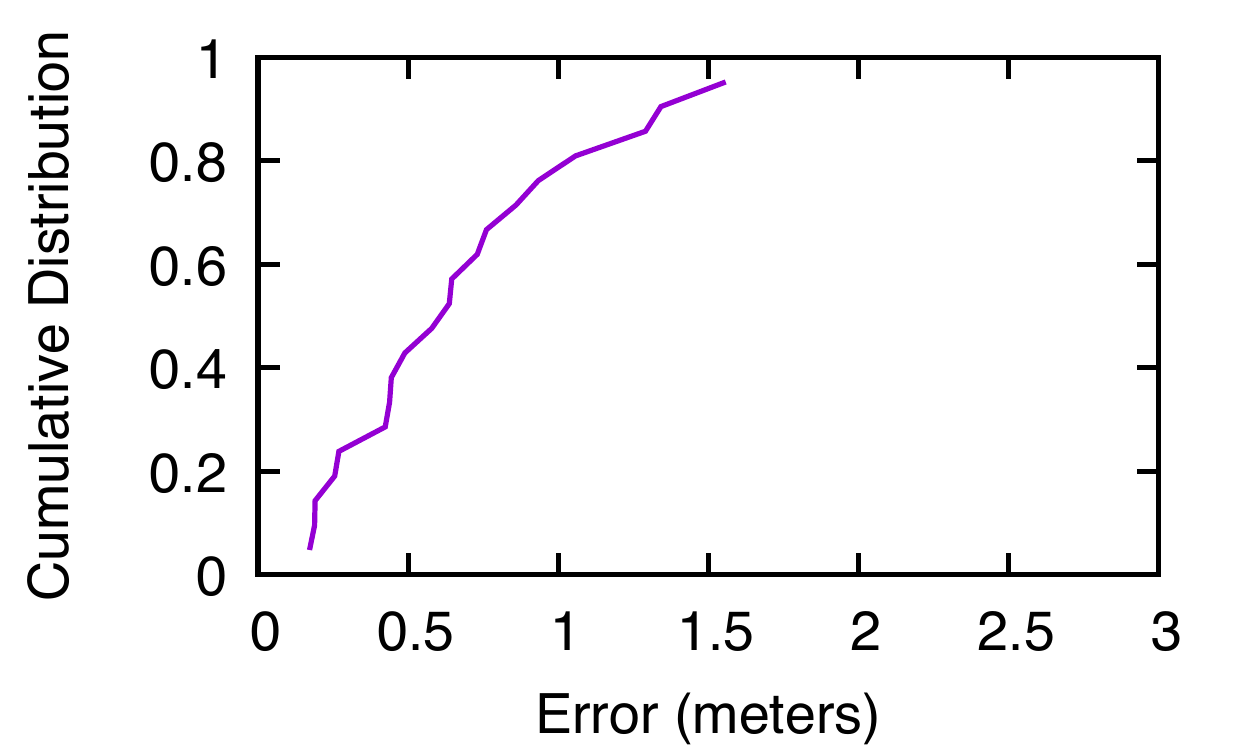}
    \caption{CDF of localization error in a indoor setting (intermittent).}
    \label{f:accuracy_wheel_indoor}
  \end{minipage}
  \hfill
  \begin{minipage}[b]{0.32\textwidth}
    \includegraphics[width=\textwidth]{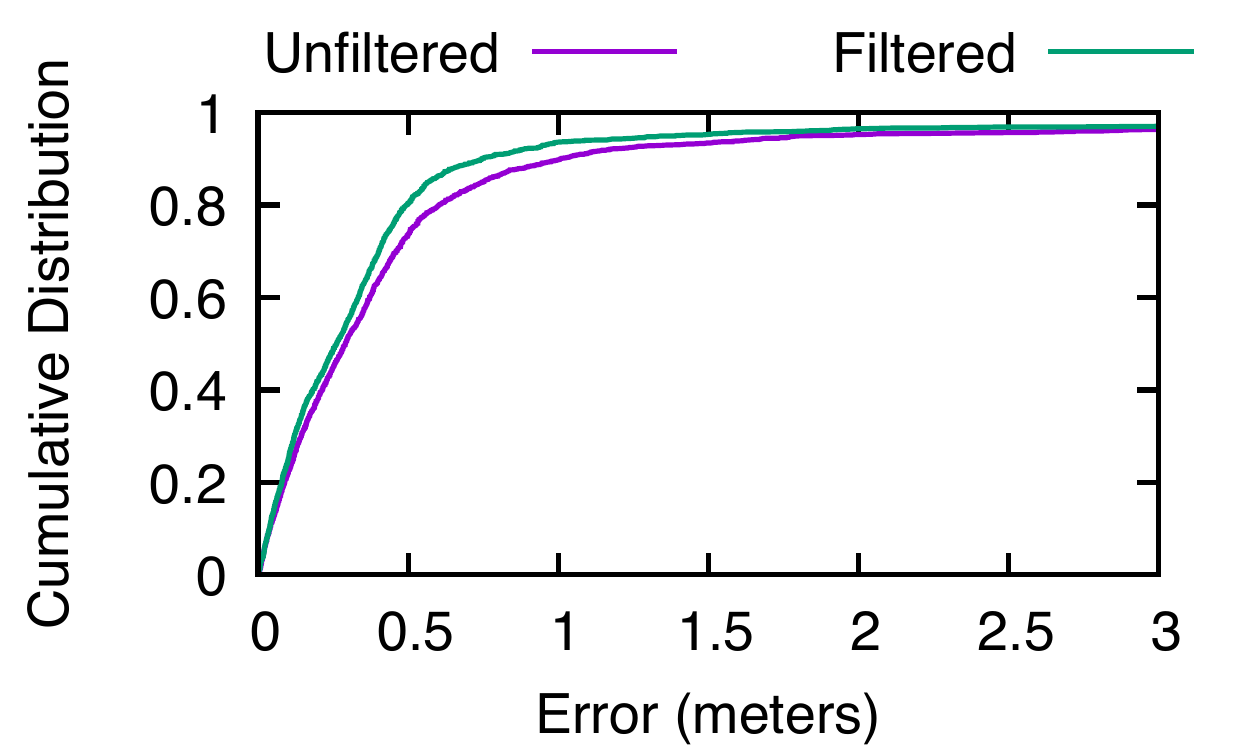}
    \caption{CDF of localization error in outdoor (continuous).}
    \label{f:accuracy_click}
  \end{minipage}
\end{figure*}

\begin{figure}
\centering
\includegraphics[width=0.45\textwidth]{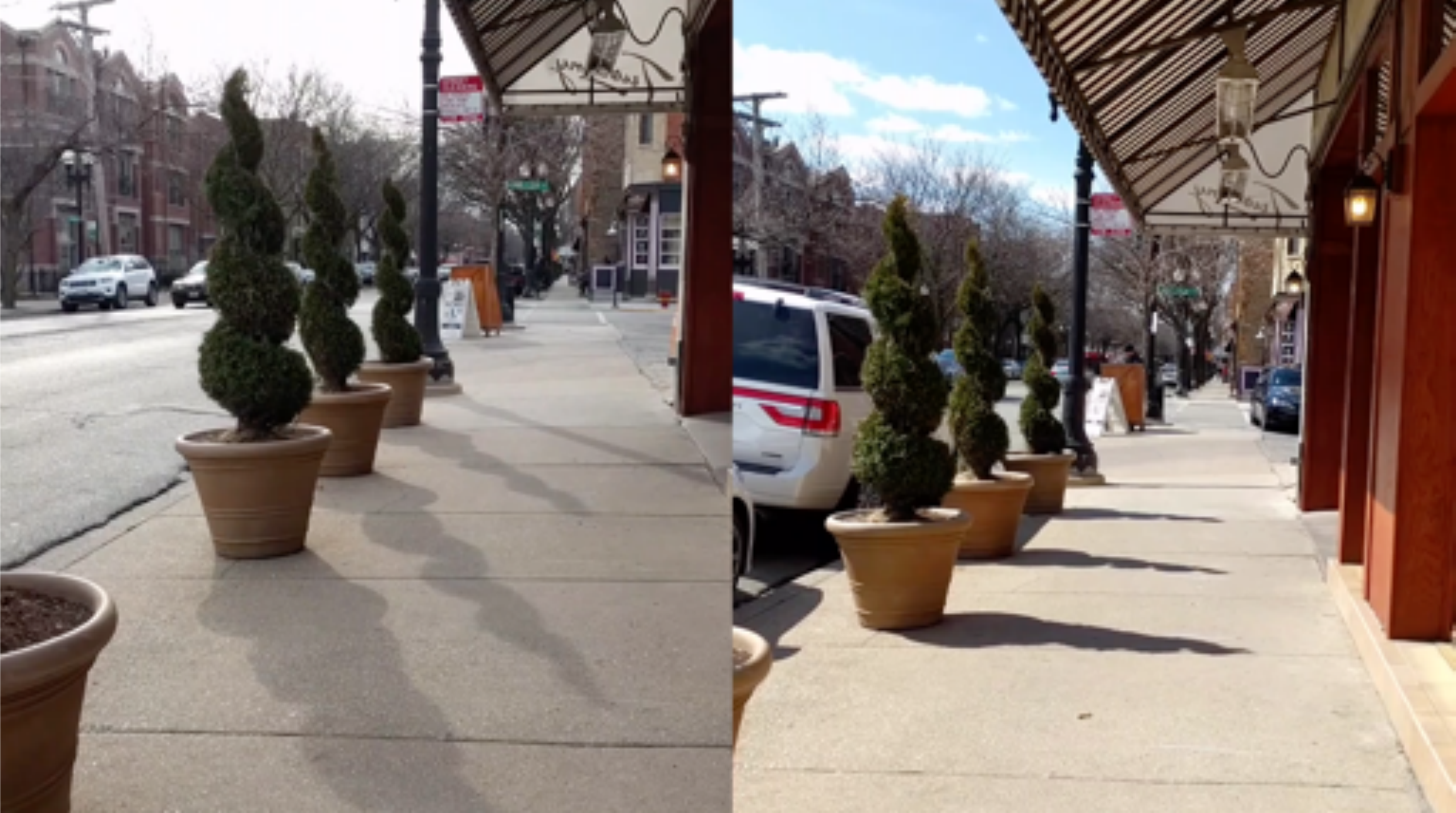}
\caption{Variation in the scene.}
\label{f:scene_variation}
\end{figure}

We collected ground truth data in both outdoors and indoors at every 5 feet as described in \S\ref{s:ground_truth_wheel}. 
In this section, we present the localization accuracy for these ground truth datasets.

Figure \ref{f:accuracy_wheel_outdoor} shows the CDF of localization error in outdoor for two different days with varying lighting condition. 
Figure \ref{f:scene_variation} shows the scene variation where the scene at the left is cloudy and the scene at the right is sunny. 
The time of day is also different as evident from the shadow.
The median localization error on these datasets is 0.35 meters for ground truth collected on the same day as the model, and 0.6 meters for data collected on a different day.

\begin{figure*}
\centering
\subfloat{\includegraphics[width=1.6in]{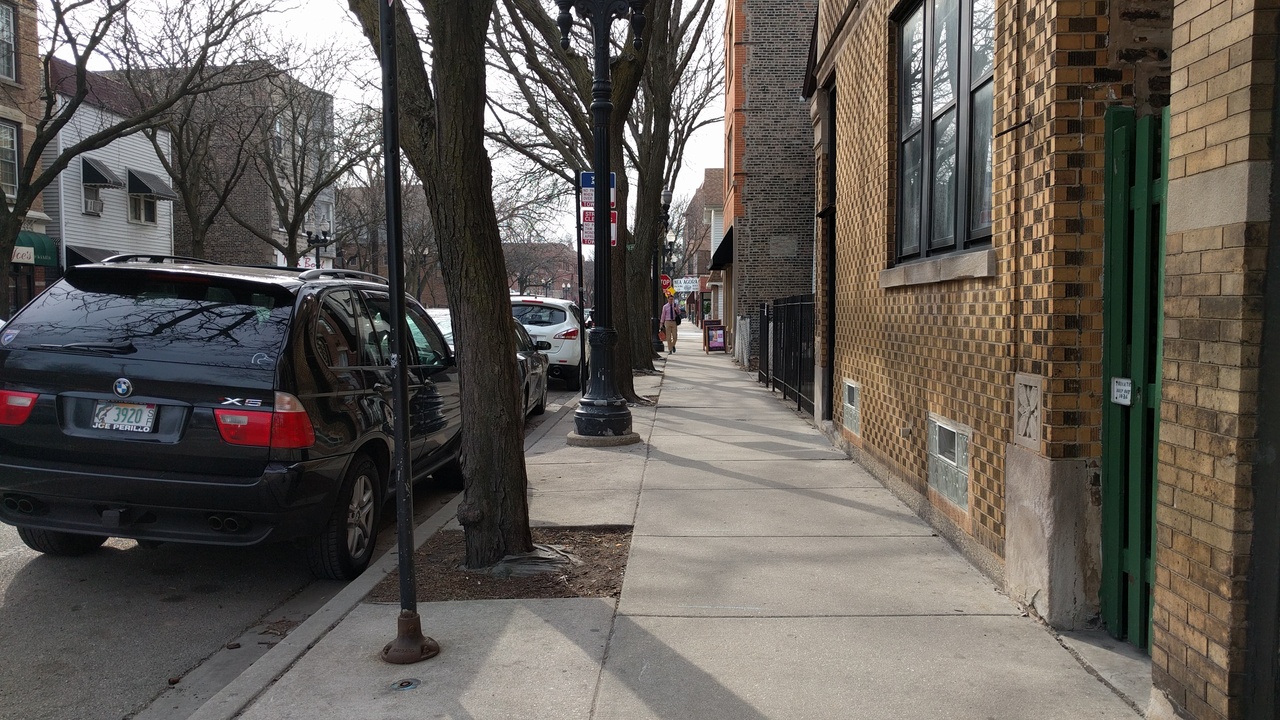}}
\vspace{0.1in}
\subfloat{\includegraphics[width=1.6in]{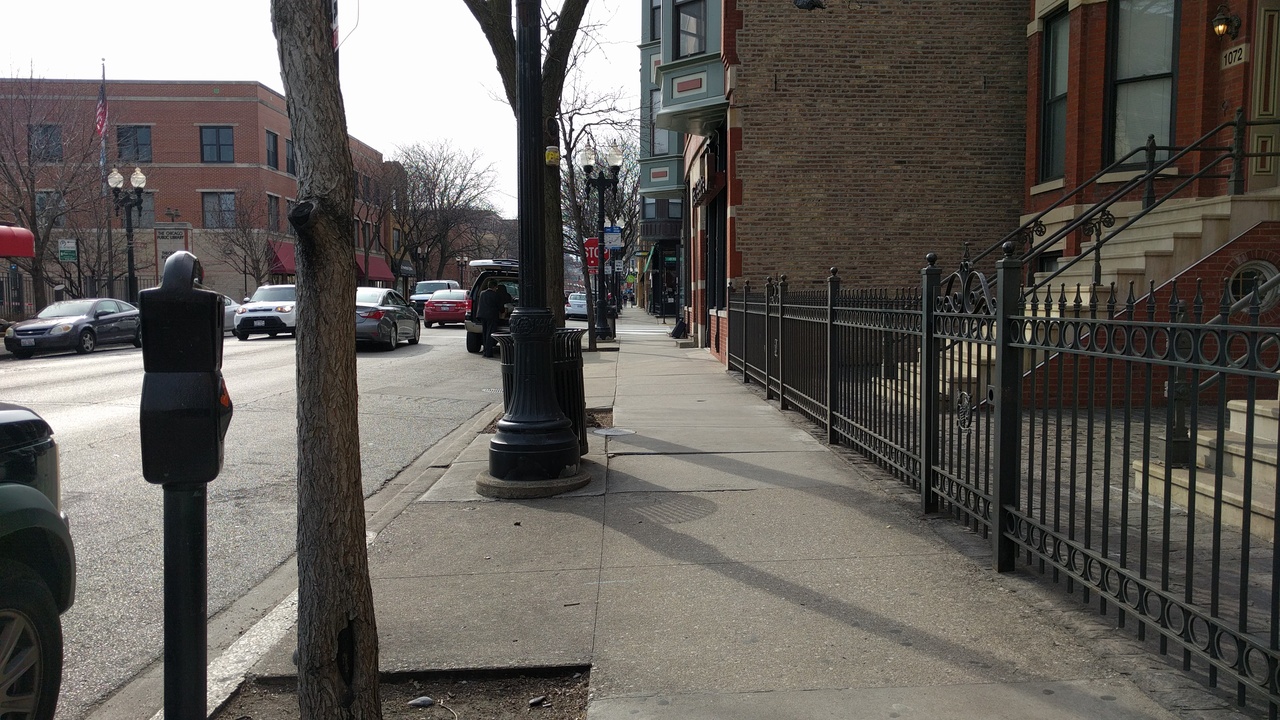}}
\vspace{0.1in}
\subfloat{\includegraphics[width=1.6in]{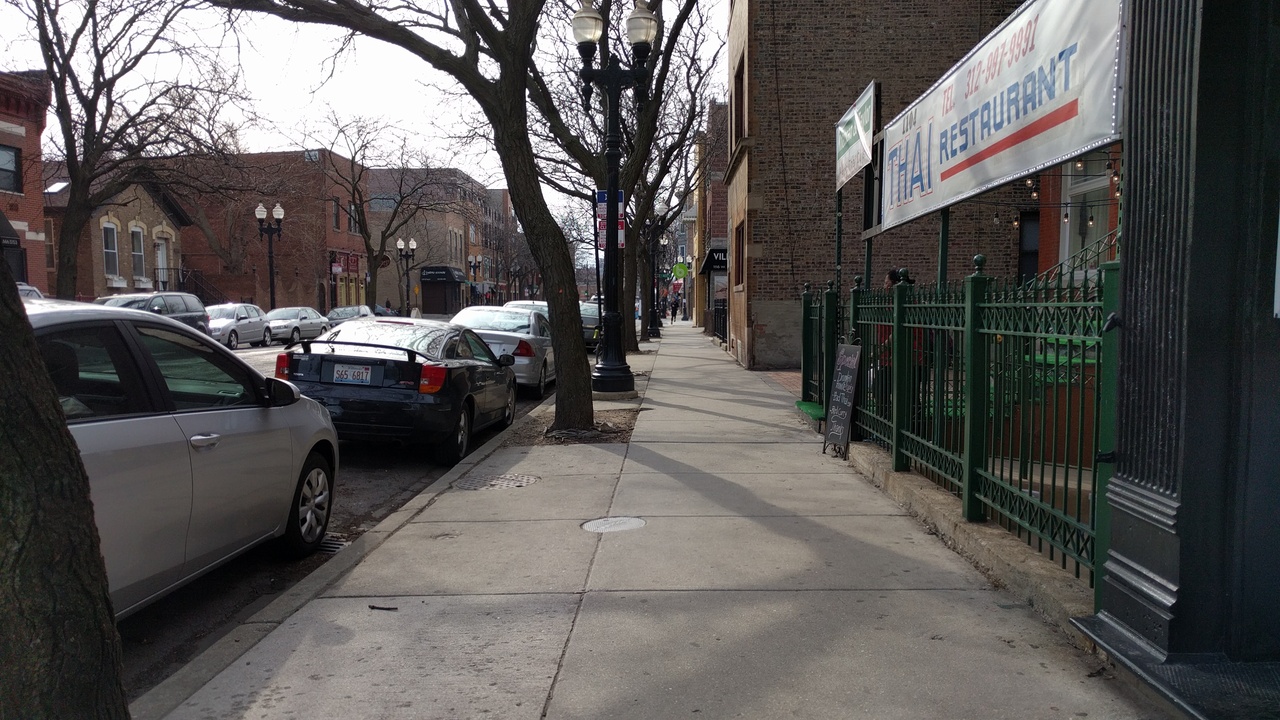}}
\vspace{0.1in}
\subfloat{\includegraphics[width=1.6in]{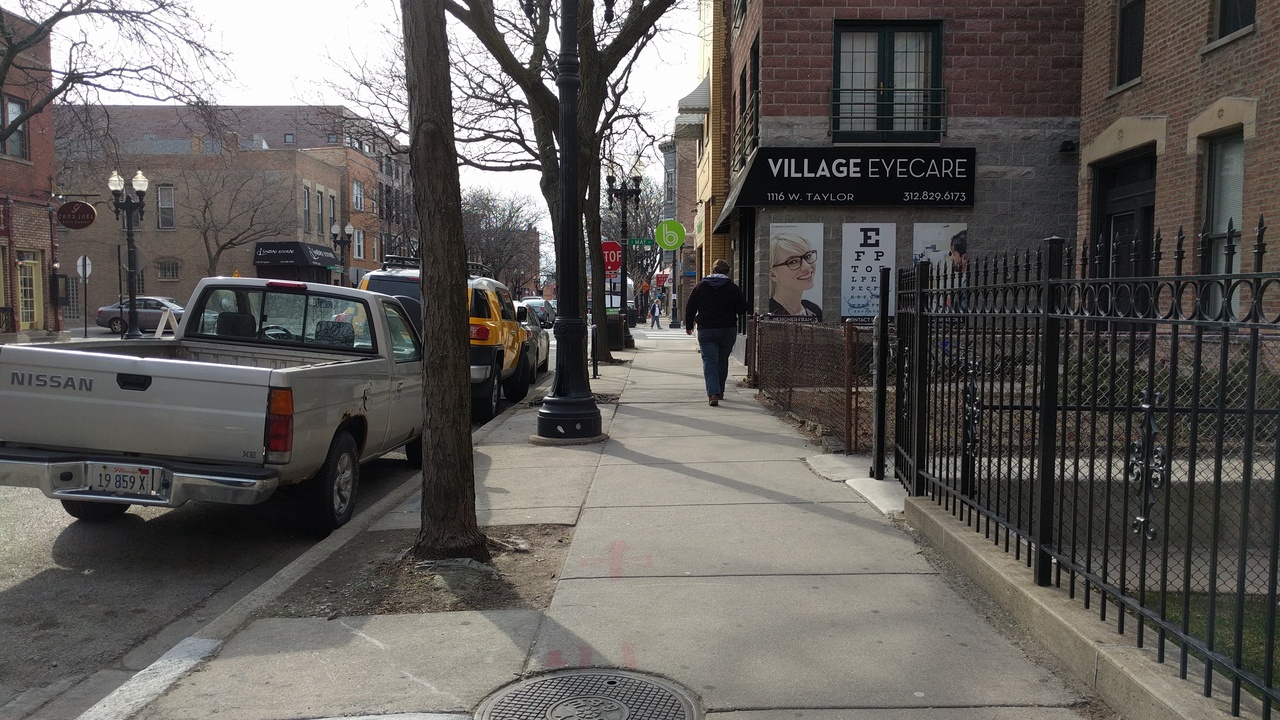}}
\vspace{-0.3in}
\caption{Examples of repetitive brick texture in wall.}
\label{f:textures}
\end{figure*}

It is worth noting that approximately 15\% locations in Figure \ref{f:accuracy_wheel_outdoor} for the different day has error higher than 5 meters. According to our investigation, the primary reason behind this is repetitive brick textures in the scene. Figure \ref{f:textures} shows some examples of this. The Kalman filter removes many of these spurious cases, which we discuss in the next section.

Figure \ref{f:accuracy_wheel_indoor} shows the CDF of localization error in a indoor setting.
Since the lighting condition generally remains the same in the indoor setting, we collected one ground truth dataset.
Here, the median error is 0.6 meter.



\subsection{Localization accuracy for continuous video}
In this section, we report the localization accuracy outdoor for ground truth data for every frame in a video sequence, which we describe in \S\ref{s:ground_truth_click}.
Figure \ref{f:accuracy_click} shows the CDF of localization error. Here, the median error is 0.3 meters for locations estimated before Kalman filtering. The median error reduces further with Kalman filtering. Note that Kalman filtering is not applicable for the intermittent dataset described above since a Kalman filter needs continuous video frames.

We note that the error has a long tail distribution, with 2\% of errors above 3 meters, and occasional much larger errors (up to 40 meters in these experiments).
These high errors occur primarily due to repetitive textures as discussed in the section above.
Large and abrupt pose changes may be eliminated through inertial sensor fusion \cite{visual-inertial}, which falls outside the scope of this paper



\subsection{Effect of compression}
\label{s:compression_effect}

\begin{figure}[ht]
  \centering
  \begin{minipage}[b]{0.45\textwidth}
    \includegraphics[width=\textwidth]{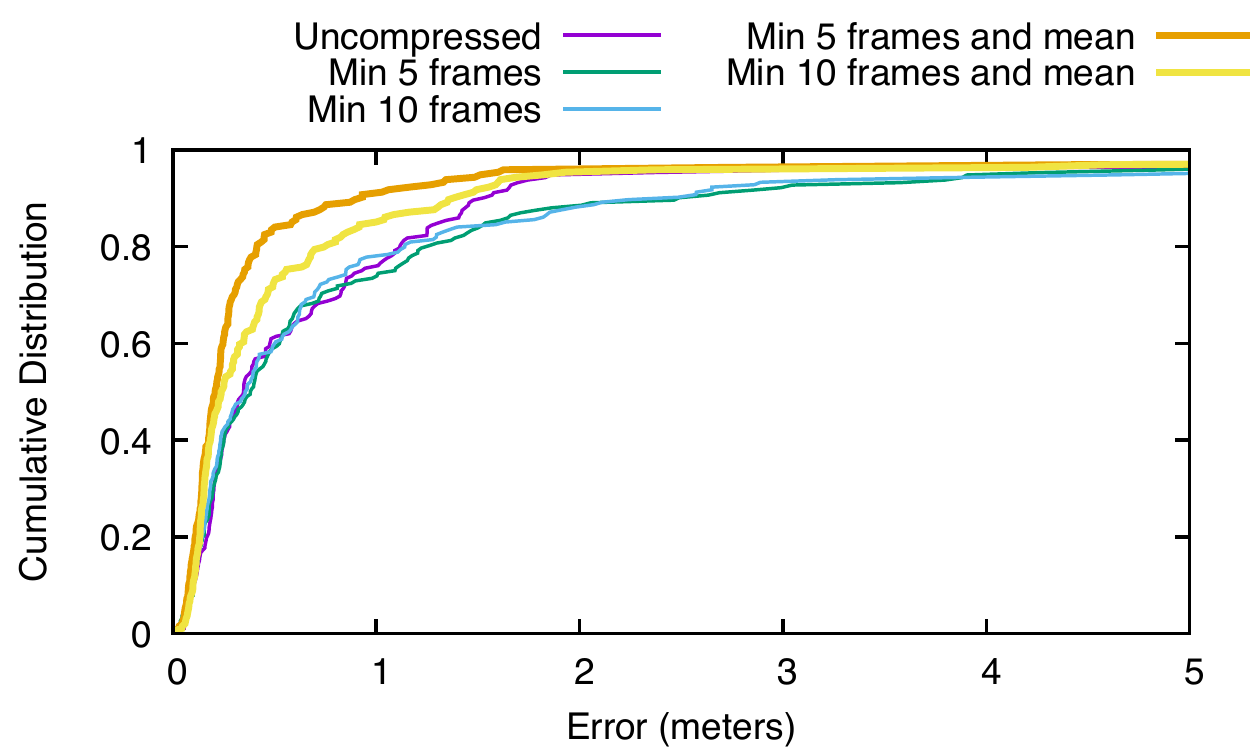}
    \caption{CDF of localization error with uncompressed 3D model, compressed 3D model with points having minimum 5 and 10 frames, as well as with mean of the descriptors (same day).}
    \label{f:compression_sameday}
  \end{minipage}
  \hfill
  \begin{minipage}[b]{0.45\textwidth}
    \includegraphics[width=\textwidth]{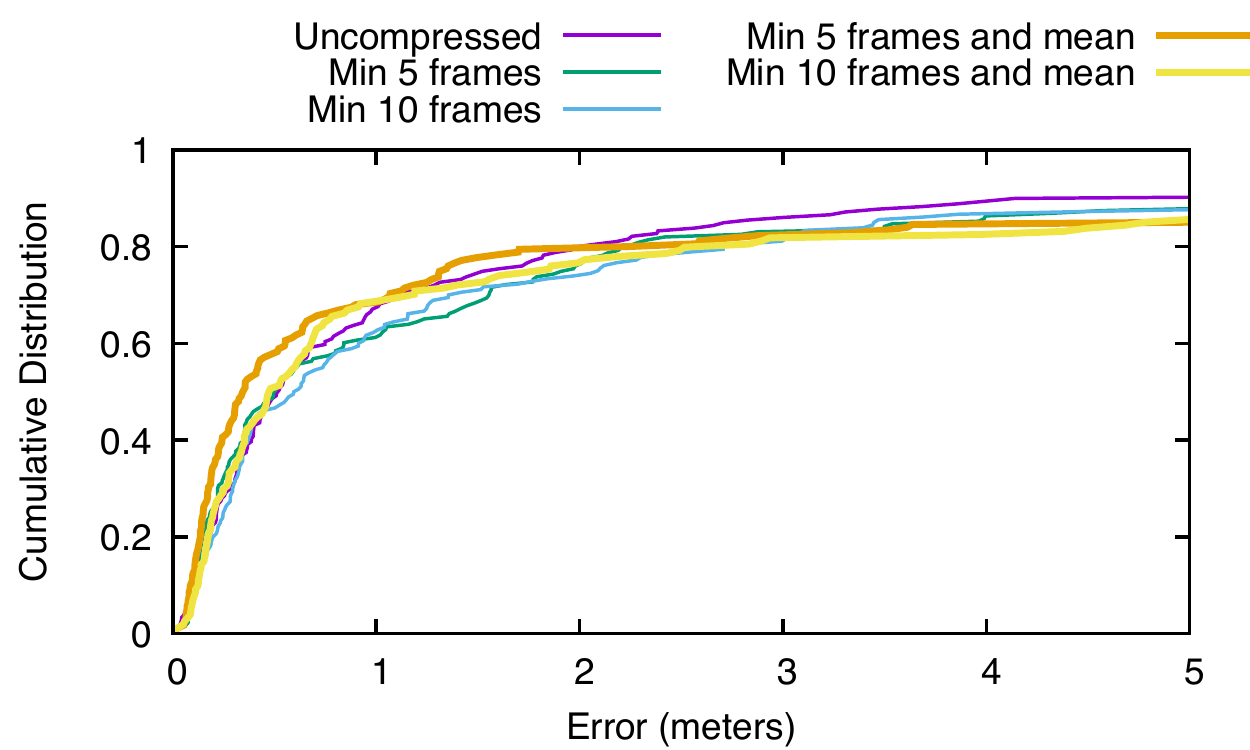}
    \caption{CDF of localization error with uncompressed 3D model, compressed 3D model with points having minimum 5 and 10 frames, as well as with mean of the descriptors (different day).}
    \label{f:compression_diffday}
  \end{minipage}
\end{figure}

Figure \ref{f:compression_sameday} and Figure \ref{f:compression_diffday} show the localization error CDFs for a uncompressed 3D model and compressed 3D models with various settings for same day and different day.
Specifically, the compression settings are (a) point reduction, and (b) mean of descriptors along with point reduction (discussed in \S\ref{s:3d_model_compression}).

Overall, Figure \ref{f:compression_sameday} and Figure \ref{f:compression_diffday} show that compression of 3D model does not reduce the localization accuracy significantly. 
Additionally, compression of a 3D model by removing points that have a low number of frame correspondences can sometimes improve accuracy as seen from Figure \ref{f:compression_sameday} for errors below 1 meter and from Figure \ref{f:compression_diffday} for errors below 0.5 meters.
This happens as pose estimation can be more accurate if we discard bad 3D points.
Surprisingly, model compression using descriptor means also improves the localization accuracy in some cases. 
We suspect that using the descriptor mean results in lower feature matching ambiguity, but we did not investigate this further.

\subsection{Smartphone computation time}

Table \ref{tab:time} shows the computation time of primary stages of the localization pipeline for both a server and a smartphone (Nexus 5X). 
We did not implement the descriptor matching and pose estimation for the Android or iOS platform because of problems with porting the relevant libraries.
However, we estimated the approximate computation time on the Nexus 5X smartphone by comparing the relative performance between the smartphone and the server.
Note that these numbers are a rough estimation and can vary in a real smartphone implementation.
We implemented keypoint tracking on the smartphone and report the computation time obtained directly from the smartphone at Table \ref{tab:time}. 

The descriptor matching stage runs in the background with additional threads besides the main thread. 
For each location estimation, we need to use keypoint tracking and pose estimation. 
These two stages take an estimated 0.25 seconds combined on a smartphone. 
Hence, we can achieve 4 location estimations per seconds, which is sufficient for effective real-time localization on a smartphone.
Note that most GPS receivers provide locations at a one second interval, and some higher-end receivers compute 5 locations every second.
Also, additional performance gains may be available through the use of GPU computing - all results presented here are from pure CPU implementations.

\begin{table}
\centering
\begin{tabular}{lll}
Task & Server & Smartphone \\
\hline
Descriptor Matching  &  $0.02$ & $0.2^*$ \\
Keypoint Tracking  & $0.011$ & $0.11$ \\
Pose Estimation  &   $0.014$ & $0.14^*$ \\ 
\hline
\end{tabular}
\caption{Computation time at server and smartphone (in seconds). The descriptor or correspondence count here is 300. ($^*$estimated as 10$\times$ server time, based on independent measurements).}
\label{tab:time}
\end{table}

\section{Related Work}
\label{s:related_visloc}

Localization by matching an image with a database of reference images is similar to image retrieval techniques. 
Brute-force matching is very expensive. 
Hence approximate matching is used, where test image descriptors are searched and matched with database descriptors using Approximate Nearest Neighbor (ANN) algorithms.
One popular ANN method is the use of KD-tree \cite{kd-tree-search1,kd-tree-search2}. However, KD-tree only builds an index with the database descriptors and all original descriptors are required during searching. 
Although KD-tree makes the searching fast, the storage requirement remains high. Hence KD-tree is typically used for only small databases.
To alleviate the storage problem, most of the recent work use vocabulary tree of visual words created with hierarchical K-means clustering \cite{vocabulary-tree,video-google}, where many descriptors are merged into a single cluster center, thus reducing the storage requirement.

In vocabulary tree, retrieval performance decreases with the larger reference database. 
Reducing the number of descriptors in the database is important for both improving retrieval performance and reducing storage requirement.
In \cite{city-scale-location}, information gain is used to keep only important descriptors in the reference database, and this enabled them to support 10 times more images in the database. 
A simple but effective technique is used in \cite{confusing-features} to remove confusing features such as those found on trees, streets. 
For each query image, they simply find the top K reference images that are not from the same location using vocabulary tree matching. If some parts of the query image match well with these retrieved images then those parts are marked as confusing.
In \cite{predicting-matchability}, a simple random decision forest classifier is trained to predict matchable descriptors. They reduced the descriptors to 30\% of the original while keeping 60\% of matchable descriptors. 

Simultaneous Localization And Mapping (SLAM) is another overlapping area of work for image-based localization.
SLAM \cite{cd_slam,orb_slam,slam_robot} constructs a map from a sequence of frames while simultaneously localizing all frames with respect to that map. 
However, SLAM encounters drift and accumulates error because of dead-reckoning. 
The primary way of detecting and correcting the accumulated error is loop-closing \cite{loop-closing-survey}, where the previously visited places are identified. 
Image matching is used for identifying the same location. 
Since brute-force matching is computationally prohibitive, approximate matching techniques \cite{dbow,fabmap} are used in practice. 
One major drawback with SLAM techniques is that the localization is only with respect to the locally constructed map, and global localization is not considered except for the loop closing.
Additionally, SLAM uses a sequence of frames or video where scene conditions (lighting, obstacles, etc.) are constant. Localization with previously captured images is more challenging due to varying scene conditions. 

Recently, the 3D model constructed using Structure-from-motion (SfM) is being increasingly used for image-based localization.
This is a very attractive method of localization as it provides 6-DOF pose instead of simple closest matching reference image provided by the image retrieval techniques discussed earlier.
Here, both fast matching of an image with respect to the reference 3D model and reducing the size of the 3D model is important for efficiency.
Unlike 2D features, a nice property of the 3D model is that it contains structural information, which can be used for both fast searching and 3D model compression.

For fast searching, \cite{prioritized-matching} used the structural property of the 3D model to prioritize the search. 
\cite{2d-3d-match} performed 2D-to-3D matching using a vocabulary tree of descriptors along with corresponding 3D points associated with each visual word. During matching, they go by decreasing order of total associated 3D points for the descriptors. 
In \cite{2d-3d-and-3d-2d}, both 2D-to-3D matching and 3D-to-2D matching are used. 
Since there are much more 2D descriptors than the corresponding 3D points, 3D-to-2D matching is more efficient. 
However, the matching quality tends to be lower in 3D-to-2D matching compared to 2D-to-3D matching where the ratio test can eliminate bad quality matches.
\cite{2d-2d-match} used efficient 2D-to-2D matching that also retrieves corresponding 3D points for pose estimation.

Compression of 3D model reduces the storage requirement and also makes the searching faster. 
\cite{prioritized-matching} formulated the model compression as a set-cover problem where a reduced point set is selected such that they are visible at a minimum of N images. 
While \cite{prioritized-matching} used simple greedy set-cover formulation to reduce the 3D-points, \cite{quadratic-reduction} used mixed-integer quadratic programming for 3D point reduction. This resulted in more flexibility and parameterization such as weights on the 3D points, occurrence, and co-occurrence of 3D points, etc. 
\cite{probabilistic-reduction} used both descriptor distinctiveness and probabilistic modeling of the set cover problem to reduce 3D points by more than 98\%.
In \cite{outdoor-localization-intel}, 3D points are reduced by removing points that do not belong to planes and lines. This essentially reduces the 3D points while keeping the 3D structure preserved.

Large-scale image-based localization is being attempted recently because of the availability of a lot of imagery data. 
Hence, fast searching and reduction of the 3D model are becoming even more important.
In \cite{world-wide-pose}, worldwide localization is attempted for two million images and 70 million reconstructed 3D points.
They could localize successfully even for 1\% inliers by introducing prior in the RANSAC selection step.

Real-time localization is still challenging because of slow descriptor computation and searching.
In \cite{real-time-msr}, real-time localization is attempted with a combination of tracking and matching. Tracking is fast, and expensive matching with real-valued features is only used when a sufficient number of points could not be tracked. 
They used smartphones' local memory to store the image descriptors, the 3D point cloud map, and various indices. 
However, for large reference map, the local storage can become infeasible. 
In contrast, \cite{mobile-server} only uses a small local map in the mobile and perform local tracking and localization, and simultaneously keeps the global map at the server. 
Then they perform alignment of a local and global map to find the 6-DOF pose of a camera with respect to the server side global map. 
However, the trade-off here is that this method encounters significant bandwidth cost.
\cite{get-out-of-my-lab} focuses on compressing the reference dataset by both 3D model reduction using set-cover formulation and descriptor quantization for real-time localization. During localization, they use local SLAM along with descriptor matching and IMU data. In contrast, our system uses only visual features and interleaved tracking and matching for fast localization.

Machine learning and Convolutional Neural Network (CNN) are successful and popular for computer vision applications nowadays.
Recently there has been work \cite{posenet, geometric-pose, lstm-pose, geometry-aware-pose} on 6-DOF pose estimation based on learning and CNN techniques.
These methods offer end-to-end learning and often learn high-level features in addition to corner features. 
However, the primary constraint in learning based methods is the training requirement.
Unlike feature-based techniques, these systems often perform poorly if the test data significantly defer from the training data.
Additionally, these techniques often require hyper-parameter tuning for each training set separately.

\section{Conclusion}
\label{s:concl}

In conclusion, we present a system to achieve sub-meter accurate localization using video analysis.
Consumer grade GPS receivers in smartphones generally encounter 5-10 meters \cite{gps_accuracy1} of error under ideal conditions, and significantly more under non-ideal conditions, such as in urban canyons.
We demonstrated that localization based on video analysis is feasible despite compute-intensive video processing by careful design and optimization of various stages of the pipeline. 
We also demonstrated that sub-meter localization accuracy can be achieved in both indoors and outdoors.

\bibliographystyle{abbrv}
\bibliography{ref}

\begin{thebibliography}{10}

\bibitem{kalman_model}
http://campar.in.tum.de/chair/kalmanfilter.

\bibitem{kalman-tutorial}
https://github.com/rlabbe/kalman-and-bayesian-filters-in-python.

\bibitem{bundler_url}
http://www.cs.cornell.edu/~snavely/bundler.

\bibitem{sfm-rome}
S.~Agarwal, Y.~Furukawa, N.~Snavely, I.~Simon, B.~Curless, S.~M. Seitz, and
  R.~Szeliski.
\newblock Building rome in a day.
\newblock {\em Commun. ACM}, 54(10):105--112, Oct. 2011.

\bibitem{wide_area_loc}
C.~Arth, D.~Wagner, M.~Klopschitz, A.~Irschara, and D.~Schmalstieg.
\newblock Wide area localization on mobile phones.
\newblock In {\em 2009 8th IEEE International Symposium on Mixed and Augmented
  Reality}, pages 73--82, Oct 2009.

\bibitem{sorroundsense}
M.~Azizyan, I.~Constandache, and R.~Roy~Choudhury.
\newblock Surroundsense: Mobile phone localization via ambience fingerprinting.
\newblock In {\em Proceedings of the 15th Annual International Conference on
  Mobile Computing and Networking}, MobiCom '09, pages 261--272, New York, NY,
  USA, 2009. ACM.

\bibitem{radar}
P.~Bahl and V.~Padmanabhan.
\newblock Radar: an in-building rf-based user location and tracking system.
\newblock In {\em Nineteenth Annual Joint Conference of the IEEE Computer and
  Communications Societies. Proceedings. IEEE (INFOCOM '00)}, volume~2, 2000.

\bibitem{kd-tree-search1}
J.~L. Bentley.
\newblock Multidimensional binary search trees used for associative searching.
\newblock {\em Commun. ACM}, 18(9):509--517, Sept. 1975.

\bibitem{walrus}
G.~Borriello, A.~Liu, T.~Offer, C.~Palistrant, and R.~Sharp.
\newblock Walrus: Wireless acoustic location with room-level resolution using
  ultrasound.
\newblock In {\em Proceedings of the 3rd International Conference on Mobile
  Systems, Applications, and Services}, MobiSys '05, pages 191--203, New York,
  NY, USA, 2005. ACM.

\bibitem{geometry-aware-pose}
S.~Brahmbhatt, J.~Gu, K.~Kim, J.~Hays, and J.~Kautz.
\newblock Geometry-aware learning of maps for camera localization.
\newblock In {\em 2018 {IEEE} Conference on Computer Vision and Pattern
  Recognition, {CVPR} 2018, Salt Lake City, UT, USA, June 18-22, 2018}, pages
  2616--2625, 2018.

\bibitem{probabilistic-reduction}
S.~Cao and N.~Snavely.
\newblock Minimal scene descriptions from structure from motion models.
\newblock In {\em Computer Vision and Pattern Recognition (CVPR), 2014 IEEE
  Conference on}, pages 461--468, June 2014.

\bibitem{loc_fine_grained}
D.~Chen, L.~Du, Z.~Jiang, W.~Xi, J.~Han, K.~Zhao, J.~Zhao, Z.~Wang, and R.~Li.
\newblock A fine-grained indoor localization using multidimensional wi-fi
  fingerprinting.
\newblock In {\em 2014 20th IEEE International Conference on Parallel and
  Distributed Systems (ICPADS)}, pages 494--501, Dec 2014.

\bibitem{outdoor-localization-intel}
K.~Chen, C.~Wang, X.~Wei, Q.~Liang, M.~Yang, C.~Chen, and Y.~Hung.
\newblock To know where we are: Vision-based positioning in outdoor
  environments.
\newblock {\em CoRR}, abs/1506.05870, 2015.

\bibitem{metropolitan-wifi}
Y.-C. Cheng, Y.~Chawathe, A.~LaMarca, and J.~Krumm.
\newblock Accuracy characterization for metropolitan-scale wi-fi localization.
\newblock In {\em Proceedings of the 3rd International Conference on Mobile
  Systems, Applications, and Services}, MobiSys '05, pages 233--245, New York,
  NY, USA, 2005. ACM.

\bibitem{loc_wo_pain}
K.~Chintalapudi, A.~Padmanabha~Iyer, and V.~N. Padmanabhan.
\newblock Indoor localization without the pain.
\newblock In {\em Proceedings of the Sixteenth Annual International Conference
  on Mobile Computing and Networking}, MobiCom '10, pages 173--184, New York,
  NY, USA, 2010. ACM.

\bibitem{ransac_eval}
S.~Choi, T.~Kim, and W.~Yu.
\newblock Performance evaluation of ransac family.
\newblock In {\em BMVC}, 2009.

\bibitem{optimal_random_ransac}
O.~Chum and J.~Matas.
\newblock Optimal randomized ransac.
\newblock {\em IEEE Transactions on Pattern Analysis and Machine Intelligence},
  30(8):1472--1482, Aug 2008.

\bibitem{placelab}
I.~S. et. al.
\newblock Place lab: Device positioning using radio beacons in the wild.
\newblock In {\em In Proceedings of the Third International Conference on
  Pervasive Computing (Pervasive '05)}. Springer, 2005.

\bibitem{ransac}
M.~A. Fischler and R.~C. Bolles.
\newblock Random sample consensus: A paradigm for model fitting with
  applications to image analysis and automated cartography.
\newblock {\em Commun. ACM}, 24(6):381--395, June 1981.

\bibitem{kd-tree-search2}
J.~H. Friedman, J.~L. Bentley, and R.~A. Finkel.
\newblock An algorithm for finding best matches in logarithmic expected time.
\newblock {\em ACM Trans. Math. Softw.}, 3(3):209--226, Sept. 1977.

\bibitem{dbow}
D.~Galvez-López and J.~D. Tardos.
\newblock Bags of binary words for fast place recognition in image sequences.
\newblock {\em IEEE Transactions on Robotics}, 28(5):1188--1197, Oct 2012.

\bibitem{p3p}
X.-S. Gao, X.-R. Hou, J.~Tang, and H.-F. Cheng.
\newblock Complete solution classification for the perspective-three-point
  problem.
\newblock {\em IEEE Transactions on Pattern Analysis and Machine Intelligence},
  25(8):930--943, Aug 2003.

\bibitem{kitti-autonomous}
A.~Geiger, P.~Lenz, and R.~Urtasun.
\newblock Are we ready for autonomous driving? the kitti vision benchmark
  suite.
\newblock In {\em Conference on Computer Vision and Pattern Recognition
  (CVPR)}, 2012.

\bibitem{virtualtourist}
T.~Goedem{\'e}, B.~Fasel, and L.~Van~Gool.
\newblock The visual virtual tourist guide: a markerless camera-based lbs
  system.
\newblock 2006.

\bibitem{active_bat}
A.~Harter, A.~Hopper, P.~Steggles, A.~Ward, and P.~Webster.
\newblock The anatomy of a context-aware application.
\newblock {\em Wireless Networks}, 8(2):187--197, 2002.

\bibitem{predicting-matchability}
W.~Hartmann, M.~Havlena, and K.~Schindler.
\newblock Predicting matchability.
\newblock In {\em Computer Vision and Pattern Recognition (CVPR), 2014 IEEE
  Conference on}, pages 9--16, June 2014.

\bibitem{loc_light}
P.~Hu, L.~Li, C.~Peng, G.~Shen, and F.~Zhao.
\newblock Pharos: Enable physical analytics through visible light based indoor
  localization.
\newblock In {\em Proceedings of the Twelfth ACM Workshop on Hot Topics in
  Networks}, HotNets-XII, pages 5:1--5:7, New York, NY, USA, 2013. ACM.

\bibitem{apolloscape}
X.~Huang, P.~Wang, X.~Cheng, D.~Zhou, Q.~Geng, and R.~Yang.
\newblock The apolloscape open dataset for autonomous driving and its
  application.
\newblock {\em ArXiv e-prints}, 2018.

\bibitem{loc_sfm}
A.~Irschara, C.~Zach, J.~M. Frahm, and H.~Bischof.
\newblock From structure-from-motion point clouds to fast location recognition.
\newblock In {\em 2009 IEEE Conference on Computer Vision and Pattern
  Recognition}, pages 2599--2606, June 2009.

\bibitem{2d-2d-match}
A.~Irschara, C.~Zach, J.-M. Frahm, and H.~Bischof.
\newblock From structure-from-motion point clouds to fast location recognition.
\newblock In {\em Computer Vision and Pattern Recognition, 2009. CVPR 2009.
  IEEE Conference on}, pages 2599--2606, June 2009.

\bibitem{fabmap}
M.~Joseph~Cummins and P.~M.~Newman.
\newblock Fab-map: Probabilistic localization and mapping in the space of
  appearance.
\newblock {\em I. J. Robotic Res.}, 27:647--665, 06 2008.

\bibitem{geometric-pose}
A.~Kendall and R.~Cipolla.
\newblock Geometric loss functions for camera pose regression with deep
  learning.
\newblock pages 6555--6564, 07 2017.

\bibitem{posenet}
A.~Kendall, M.~Grimes, and R.~Cipolla.
\newblock Posenet: A convolutional network for real-time 6-dof camera
  relocalization.
\newblock In {\em ICCV}, pages 2938--2946. IEEE Computer Society, 2015.

\bibitem{confusing-features}
J.~Knopp, J.~Sivic, and T.~Pajdla.
\newblock Avoiding confusing features in place recognition.
\newblock In {\em Proceedings of the 11th European Conference on Computer
  Vision: Part I}, ECCV'10, pages 748--761, Berlin, Heidelberg, 2010.
  Springer-Verlag.

\bibitem{zero_cost}
S.~Kumar, S.~Gil, D.~Katabi, and D.~Rus.
\newblock Accurate indoor localization with zero start-up cost.
\newblock In {\em Proceedings of the 20th Annual International Conference on
  Mobile Computing and Networking}, MobiCom '14, pages 483--494, New York, NY,
  USA, 2014. ACM.

\bibitem{slam_robot}
J.~J. Leonard and H.~F. Durrant-Whyte.
\newblock Simultaneous map building and localization for an autonomous mobile
  robot.
\newblock In {\em Intelligent Robots and Systems '91. 'Intelligence for
  Mechanical Systems, Proceedings IROS '91. IEEE/RSJ International Workshop
  on}, pages 1442--1447 vol.3, Nov 1991.

\bibitem{epnp}
V.~Lepetit, F.~Moreno-Noguer, and P.~Fua.
\newblock Epnp: An accurate o(n) solution to the pnp problem.
\newblock {\em International Journal of Computer Vision}, 81(2):155, 2008.

\bibitem{world-wide-pose}
Y.~Li, N.~Snavely, D.~Huttenlocher, and P.~Fua.
\newblock Worldwide pose estimation using 3d point clouds.
\newblock In {\em Proceedings of the 12th European Conference on Computer
  Vision - Volume Part I}, ECCV'12, pages 15--29, Berlin, Heidelberg, 2012.
  Springer-Verlag.

\bibitem{prioritized-matching}
Y.~Li, N.~Snavely, and D.~P. Huttenlocher.
\newblock Location recognition using prioritized feature matching.
\newblock In {\em Proceedings of the 11th European Conference on Computer
  Vision: Part II}, ECCV'10, pages 791--804, Berlin, Heidelberg, 2010.
  Springer-Verlag.

\bibitem{loc_limit}
H.~Liu, Y.~Gan, J.~Yang, S.~Sidhom, Y.~Wang, Y.~Chen, and F.~Ye.
\newblock Push the limit of wifi based localization for smartphones.
\newblock In {\em Proceedings of the 18th Annual International Conference on
  Mobile Computing and Networking}, Mobicom '12, pages 305--316, New York, NY,
  USA, 2012. ACM.

\bibitem{acoustic}
K.~Liu, X.~Liu, and X.~Li.
\newblock Guoguo: Enabling fine-grained indoor localization via smartphone.
\newblock In {\em Proceeding of the 11th Annual International Conference on
  Mobile Systems, Applications, and Services}, MobiSys '13, pages 235--248, New
  York, NY, USA, 2013. ACM.

\bibitem{sift}
D.~G. Lowe.
\newblock Distinctive image features from scale-invariant keypoints.
\newblock {\em Int. J. Comput. Vision}, 60(2):91--110, Nov. 2004.

\bibitem{ap-fingerprinting-kalman}
H.~Lu, S.~Zhang, X.~Liu, and X.~Lin.
\newblock Vehicle tracking using particle filter in wi-fi network.
\newblock In {\em Vehicular Technology Conference Fall, 2010 IEEE 72nd (VTC
  '10)}, sept. 2010.

\bibitem{optical_flow}
B.~D. Lucas and T.~Kanade.
\newblock An iterative image registration technique with an application to
  stereo vision.
\newblock In {\em Proceedings of the 7th International Joint Conference on
  Artificial Intelligence - Volume 2}, IJCAI'81, pages 674--679, San Francisco,
  CA, USA, 1981. Morgan Kaufmann Publishers Inc.

\bibitem{get-out-of-my-lab}
S.~Lynen, T.~Sattler, M.~Bosse, J.~Hesch, M.~Pollefeys, and R.~Siegwart.
\newblock Get out of my lab: Large-scale, real-time visual-inertial
  localization.
\newblock In {\em Proceedings of Robotics: Science and Systems}, Rome, Italy,
  July 2015.

\bibitem{get_out_of_my_lab}
S.~Lynen, T.~Sattler, M.~Bosse, J.~A. Hesch, M.~Pollefeys, and R.~Y. Siegwart.
\newblock {G}et {O}ut of {M}y {L}ab: {L}arge-scale, {R}eal-{T}ime
  {V}isual-{I}nertial {L}ocalization.
\newblock In L.~E. Kavraki, D.~Hsu, and J.~Buchli, editors, {\em Robotics:
  Science and Systems XI}, pages 37--, S.l., 2015. RSS.

\bibitem{mobile-server}
S.~Middelberg, T.~Sattler, O.~Untzelmann, and L.~Kobbelt.
\newblock Scalable 6-dof localization on mobile devices.
\newblock In D.~Fleet, T.~Pajdla, B.~Schiele, and T.~Tuytelaars, editors, {\em
  Computer Vision – ECCV 2014}, volume 8690 of {\em Lecture Notes in Computer
  Science}, pages 268--283. Springer International Publishing, 2014.

\bibitem{flann3}
M.~Muja and D.~G. Lowe.
\newblock Fast approximate nearest neighbors with automatic algorithm
  configuration.
\newblock In {\em International Conference on Computer Vision Theory and
  Application VISSAPP'09)}, pages 331--340. INSTICC Press, 2009.

\bibitem{flann2}
M.~Muja and D.~G. Lowe.
\newblock Fast matching of binary features.
\newblock In {\em Computer and Robot Vision {(CRV)}}, pages 404--410, 2012.

\bibitem{flann1}
M.~Muja and D.~G. Lowe.
\newblock Scalable nearest neighbor algorithms for high dimensional data.
\newblock {\em Pattern Analysis and Machine Intelligence, IEEE Transactions
  on}, 36, 2014.

\bibitem{orb_slam}
R.~Mur-Artal, J.~M.~M. Montiel, and J.~D. Tardós.
\newblock Orb-slam: A versatile and accurate monocular slam system.
\newblock {\em IEEE Transactions on Robotics}, 31(5):1147--1163, Oct 2015.

\bibitem{vocabulary-tree}
D.~Nister and H.~Stewenius.
\newblock Scalable recognition with a vocabulary tree.
\newblock In {\em Proceedings of the 2006 IEEE Computer Society Conference on
  Computer Vision and Pattern Recognition - Volume 2}, CVPR '06, pages
  2161--2168, Washington, DC, USA, 2006. IEEE Computer Society.

\bibitem{quadratic-reduction}
H.~S. Park, Y.~Wang, E.~Nurvitadhi, J.~Hoe, Y.~Sheikh, and M.~Chen.
\newblock 3d point cloud reduction using mixed-integer quadratic programming.
\newblock In {\em Computer Vision and Pattern Recognition Workshops (CVPRW),
  2013 IEEE Conference on}, pages 229--236, June 2013.

\bibitem{cd_slam}
K.~Pirker, M.~Rüther, and H.~Bischof.
\newblock Cd slam - continuous localization and mapping in a dynamic world.
\newblock In {\em 2011 IEEE/RSJ International Conference on Intelligent Robots
  and Systems}, pages 3990--3997, Sept 2011.

\bibitem{cricket}
N.~B. Priyantha, A.~Chakraborty, and H.~Balakrishnan.
\newblock The cricket location-support system.
\newblock In {\em Proceedings of the 6th Annual International Conference on
  Mobile Computing and Networking}, MobiCom '00, pages 32--43, New York, NY,
  USA, 2000. ACM.

\bibitem{2d-3d-match}
T.~Sattler, B.~Leibe, and L.~Kobbelt.
\newblock Fast image-based localization using direct 2d-to-3d matching.
\newblock In {\em Computer Vision (ICCV), 2011 IEEE International Conference
  on}, pages 667--674, Nov 2011.

\bibitem{2d-3d-and-3d-2d}
T.~Sattler, B.~Leibe, and L.~Kobbelt.
\newblock Improving image-based localization by active correspondence search.
\newblock In {\em Proceedings of the 12th European Conference on Computer
  Vision - Volume Part I}, ECCV'12, pages 752--765, Berlin, Heidelberg, 2012.
  Springer-Verlag.

\bibitem{city-scale-location}
G.~Schindler, M.~Brown, and R.~Szeliski.
\newblock City-scale location recognition.
\newblock In {\em Computer Vision and Pattern Recognition, 2007. CVPR '07. IEEE
  Conference on}, pages 1--7, June 2007.

\bibitem{spotloc}
S.~Sen, B.~Radunovic, R.~R. Choudhury, and T.~Minka.
\newblock You are facing the mona lisa: Spot localization using phy layer
  information.
\newblock In {\em Proceedings of the 10th International Conference on Mobile
  Systems, Applications, and Services}, MobiSys '12, pages 183--196, New York,
  NY, USA, 2012. ACM.

\bibitem{real-time-msr}
S.~N. Sinha.
\newblock Real-time image-based 6-dof localization in large-scale environments.
\newblock In {\em Proceedings of the 2012 IEEE Conference on Computer Vision
  and Pattern Recognition (CVPR)}, CVPR '12, pages 1043--1050, Washington, DC,
  USA, 2012. IEEE Computer Society.

\bibitem{video-google}
J.~Sivic and A.~Zisserman.
\newblock Video google: a text retrieval approach to object matching in videos.
\newblock In {\em Proceedings Ninth IEEE International Conference on Computer
  Vision}, pages 1470--1477 vol.2, Oct 2003.

\bibitem{bundler}
N.~Snavely, S.~M. Seitz, and R.~Szeliski.
\newblock Photo tourism: Exploring photo collections in 3d.
\newblock In {\em ACM SIGGRAPH 2006 Papers}, SIGGRAPH '06, pages 835--846, New
  York, NY, USA, 2006. ACM.

\bibitem{bundler2}
N.~Snavely, S.~M. Seitz, and R.~Szeliski.
\newblock Modeling the world from internet photo collections.
\newblock {\em Int. J. Comput. Vision}, 80(2):189--210, Nov. 2008.

\bibitem{gps_accuracy_urban1}
Z.~Tan, D.~Chu, and L.~Zhong.
\newblock Vision: Cloud and crowd assistance for gps urban canyons.
\newblock In {\em Proceedings of the Fifth International Workshop on Mobile
  Cloud Computing \&\#38; Services}, MCS '14, pages 23--27, New York, NY, USA,
  2014. ACM.

\bibitem{visual-inertial}
P.~{Tanskanen}, T.~{Naegeli}, M.~{Pollefeys}, and O.~{Hilliges}.
\newblock Semi-direct ekf-based monocular visual-inertial odometry.
\newblock In {\em 2015 IEEE/RSJ International Conference on Intelligent Robots
  and Systems (IROS)}, pages 6073--6078, Sep. 2015.

\bibitem{blind-nav}
S.~Treuillet and E.~Royer.
\newblock Outdoor/indoor vision based localization for blind pedestrian
  navigation assistance.
\newblock {\em Int. J. Image Graphics}, 10:481--496, 10 2010.

\bibitem{global_slam_loc}
J.~Ventura, C.~Arth, G.~Reitmayr, and D.~Schmalstieg.
\newblock Global localization from monocular slam on a mobile phone.
\newblock {\em IEEE Transactions on Visualization and Computer Graphics},
  20(4):531--539, Apr. 2014.

\bibitem{lstm-pose}
F.~Walch, C.~Hazirbas, L.~Leal-Taix{\'e}, T.~Sattler, S.~Hilsenbeck, and
  D.~Cremers.
\newblock Image-based localization using lstms for structured feature
  correlation.
\newblock {\em 2017 IEEE International Conference on Computer Vision (ICCV)},
  pages 627--637, 2017.

\bibitem{kalman}
G.~Welch and G.~Bishop.
\newblock An introduction to the kalman filter.
\newblock Technical report, Chapel Hill, NC, USA, 1995.

\bibitem{loop-closing-survey}
B.~Williams, M.~Cummins, J.~Neira, P.~Newman, I.~Reid, and J.~Tardos.
\newblock A comparison of loop closing techniques in monocular \{SLAM\}.
\newblock {\em Robotics and Autonomous Systems}, 57(12):1188 -- 1197, 2009.
\newblock Inside Data Association.

\bibitem{cell-tower}
J.~Yang, A.~Varshavsky, H.~Liu, Y.~Chen, and M.~Gruteser.
\newblock Accuracy characterization of cell tower localization.
\newblock In {\em Proceedings of the 12th ACM International Conference on
  Ubiquitous Computing}, UbiComp '10, pages 223--226, New York, NY, USA, 2010.
  ACM.

\bibitem{gps_accuracy1}
P.~A. Zandbergen and S.~J. Barbeau.
\newblock Positional accuracy of assisted gps data from high-sensitivity
  gps-enabled mobile phones.
\newblock {\em Journal of Navigation}, 64(03):381--399, 2011.

\end{thebibliography}

\end{document}